\documentclass{article}

\usepackage{microtype}
\usepackage{graphicx}
\usepackage{subcaption}
\usepackage{booktabs} 

\usepackage{hyperref}

\usepackage[preprint]{icml2026}

\usepackage{enumitem}
\usepackage{amsmath}
\usepackage{amssymb}
\usepackage{mathtools}
\usepackage{longtable}
\usepackage{comment}
\usepackage{xspace}
\usepackage[textsize=tiny]{todonotes}
\usepackage[most]{tcolorbox}
\usepackage{makecell}
\usepackage{collcell}
\usepackage{booktabs}
\usepackage{pifont}
\usepackage{lipsum}

\usepackage[capitalize,noabbrev]{cleveref}
\usepackage{tabularx}
\usepackage{listings}
\usepackage[table]{xcolor}

\lstdefinestyle{py}{
  language=Python,
  basicstyle=\ttfamily\footnotesize,
  keywordstyle=\color{blue},
  stringstyle=\color{teal},
  commentstyle=\color{gray},
  showstringspaces=false,
  breaklines=true,
  breakatwhitespace=true,
  frame=single,
  columns=fullflexible
}

\usepackage{amsmath,amsfonts,bm}
\definecolor{green}{RGB}{36, 214, 36}
\definecolor{red}{RGB}{235, 30, 30}
\definecolor{lightredshade}{HTML}{dea9a9}
\definecolor{lightgreenshade}{HTML}{bce3bd}
\definecolor{lightblueshade}{HTML}{cacbe8}
\definecolor{MyDarkBlue}{rgb}{0,0.08,1}
\definecolor{MyDarkGreen}{rgb}{0.02,0.6,0.02}
\definecolor{MyDarkRed}{rgb}{0.8,0.02,0.02}
\definecolor{MyDarkOrange}{rgb}{0.40,0.2,0.02}
\definecolor{MyPurple}{RGB}{111,0,255}
\definecolor{MyRed}{rgb}{1.0,0.0,0.0}
\definecolor{MyGold}{rgb}{0.75,0.6,0.12}
\definecolor{MyDarkgray}{rgb}{0.66, 0.66, 0.66}

\definecolor{MyYellow}{rgb}{254, 246, 170}
\definecolor{MyBlue}{rgb}{170, 217, 251}
\definecolor{LuneBlue}{rgb}{0.11, 0.11, 0.43}
\newcommand{\colorDelta}[1]{%
  \ifnum#1>50\relax\cellcolor{green!65}+{#1}\%\else%
  \ifnum#1>40\relax\cellcolor{green!40}+{#1}\%\else%
  \ifnum#1>30\relax\cellcolor{green!30}+{#1}\%\else%
  \ifnum#1>20\relax\cellcolor{green!20}+{#1}\%\else%
  \ifnum#1>10\relax\cellcolor{green!10}+{#1}\%\else%
  \ifnum#1>0\relax\cellcolor{green!5}+{#1}\%\else%
  #1\%\fi\fi\fi\fi\fi\fi%
}
\newcommand{\greencheck}{\textcolor{green}{\ding{51}}}
\newcommand{\redcross}{\textcolor{red}{\ding{55}}}

\newcommand{\eg}{{\it e.g.},~}%
\newcommand{\ie}{{\it i.e.},~}%

\newcommand{\method}{\textsc{VeriEnv}\xspace}

\lstdefinestyle{python}{
    language=Python,
    basicstyle=\fontsize{8}{10}\ttfamily,
    keywordstyle=\color{blue},
    commentstyle=\color{gray},
    stringstyle=\color{black},
    showstringspaces=false,
    breaklines=true,
    breakindent=0pt,
    breakatwhitespace=false,
    escapeinside={(*@}{@*)}
}

\lstdefinestyle{cpp}{
    language=C++,
    basicstyle=\fontsize{8}{10}\ttfamily,
    keywordstyle=\color{blue},
    commentstyle=\color{gray},
    stringstyle=\color{green},
    showstringspaces=false,
    breaklines=true,
    breakindent=0pt,
    breakatwhitespace=false,
    escapeinside={(*@}{@*)}
}

\lstdefinestyle{plain}{
    basicstyle=\fontsize{8}{10}\ttfamily,
    keywordstyle=\color{blue},
    commentstyle=\color{gray},
    stringstyle=\color{green},
    showstringspaces=false,
    breaklines=true,
    breakatwhitespace=false,
    breakindent=0pt,
    escapeinside={(*@}{@*)}
}

\lstdefinestyle{python2}{
    language=Python,
    basicstyle=\fontsize{8}{10}\ttfamily,
    keywordstyle=\color{blue},
    commentstyle=\color{gray},
    stringstyle=\color{green},
    showstringspaces=false,
    breakatwhitespace=false,
    breaklines=true,
    breakindent=0pt,
    escapeinside={(*@}{@*)}
}

\lstdefinestyle{cpp2}{
    language=C++,
    basicstyle=\fontsize{8}{10}\ttfamily,
    keywordstyle=\color{blue},
    commentstyle=\color{gray},
    stringstyle=\color{green},
    showstringspaces=false,
    breaklines=true,
    breakindent=0pt,
    breakatwhitespace=false,
    escapeinside={(*@}{@*)}
}

\lstdefinestyle{sql}{
    language=SQL,
    basicstyle=\fontsize{8}{10}\ttfamily,
    keywordstyle=\color{blue},
    commentstyle=\color{green},
    stringstyle=\color{black},
    showstringspaces=false,
    breakatwhitespace=false,
    breaklines=true,
    breakindent=0pt,
    escapeinside={(*@}{@*)}
}

\lstdefinestyle{prompt}{
    language=Python,
    basicstyle=\fontsize{8}{10}\ttfamily,
    keywordstyle=\color{blue},
    commentstyle=\color{gray},
    stringstyle=\color{cppgreen},
    showstringspaces=false,
    breaklines=true,
    backgroundcolor=\color{bgcolor},
    keepspaces=true, 
    breakindent=0pt,
    breakatwhitespace=false,
    showspaces=false,   
    escapeinside={(*@}{@*)}
}
\lstdefinestyle{text}{
    basicstyle=\fontsize{8}{10}\ttfamily,
    showstringspaces=false,
    breaklines=true,
    backgroundcolor=\color{bgcolor},
    breakatwhitespace=false,
    breakindent=0pt,
    keepspaces=true,
    showspaces=false,   
    escapeinside={(*@}{@*)}
}

\def\eqref#1{equation~\ref{#1}}

\def\1{\bm{1}}

\DeclareMathAlphabet{\mathsfit}{\encodingdefault}{\sfdefault}{m}{sl}
\SetMathAlphabet{\mathsfit}{bold}{\encodingdefault}{\sfdefault}{bx}{n}

\definecolor{DarkGreen}{RGB}{30,130,30}
\definecolor{Red}{RGB}{235,74,74}

\usepackage[textsize=tiny]{todonotes}

\icmltitlerunning{Safe and Scalable Web Agent Learning via Recreated Websites}

\begin{document}

\twocolumn[
  \icmltitle{Safe and Scalable Web Agent Learning via Recreated Websites}



  \icmlsetsymbol{equal}{*}

  \begin{icmlauthorlist}
    \icmlauthor{Hyungjoo Chae}{gt}
    \icmlauthor{Jungsoo Park}{gt}
    \icmlauthor{Alan Ritter}{gt}
  \end{icmlauthorlist}

  \icmlaffiliation{gt}{School of Interactive Computing, Georgia Institute of Technology}

  \icmlcorrespondingauthor{Alan Ritter}{alan.ritter@gatech.edu}

  \icmlkeywords{Machine Learning, ICML}

  \vskip 0.3in
]



\printAffiliationsAndNotice{}  

\begin{abstract}
Training autonomous web agents is fundamentally limited by the environments they learn from: real-world websites are unsafe to explore, hard to reset, and rarely provide verifiable feedback.
We propose \method, a framework that treats language models as environment creators, automatically cloning real-world websites into fully executable, verifiable synthetic environments.
By exposing controlled internal access via a Python SDK, \method enables agents to self-generate tasks with deterministic, programmatically verifiable rewards, eliminating reliance on heuristic or LLM-based judges.
This design decouples agent learning from unsafe real-world interaction while enabling scalable self-evolution through environment expansion.
Through experiments on web agent benchmarks, we show that agents trained with \method generalize to unseen websites, achieve site-specific mastery through self-evolving training, and benefit from scaling the number of training environments.
Code and resources will be released at \url{https://github.com/kyle8581/VeriEnv}
 upon acceptance.
\end{abstract}

\begin{figure}[!ht]
    \centering
    \includegraphics[width=1.0\linewidth]{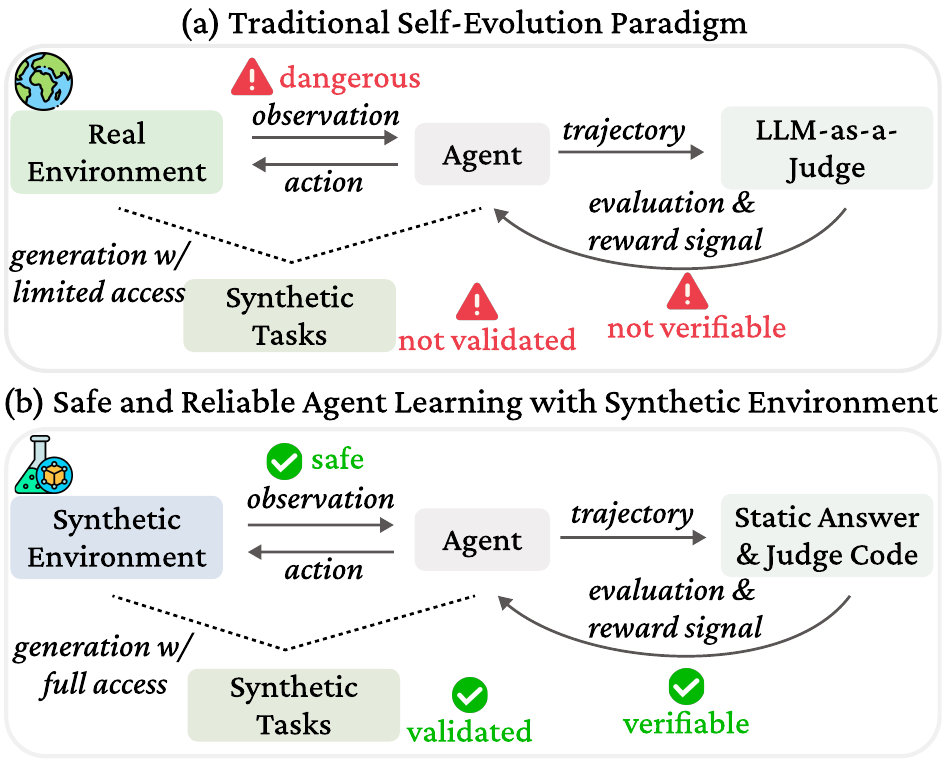}
    \caption{Comparison between the traditional self-evolution paradigm and our verifiable environment framework.
(a) In traditional settings, agents interact directly with real-world environments and rely on unvalidated synthetic tasks and non-verifiable, LLM-based reward signals, leading to unsafe exploration and unreliable learning.
(b) In contrast, \method clones real-world websites into synthetic environments with full internal access, enabling safe exploration, validated task generation, and deterministic, verifiable reward signals for stable and scalable agent learning.}
    \label{fig:motivating}
    
\end{figure}
\section{Introduction}

Autonomous computer agents that can proactively assist humans in real-world tasks are a central goal of artificial intelligence~\citep{osworld, xu2024theagentcompany}. Achieving this vision requires agents that can {self-evolve}: continuously generating new challenges, interacting with complex environments, and improving without relying on costly human data~\citep{zhou2025proposer, huang2025r}. Recent advances therefore explore reinforcement learning for web agents, where agents directly interact with real-world websites, autonomously create tasks, and learn through self-challenging paradigms~\citep{qi2025webrl}. 
Because the web constitutes one of the most realistic and diverse computer-use environments, with long-horizon interactions, rich state, and heterogeneous interfaces~\citep{zhou2024webarena, he2024webvoyager}, it provides a natural testbed for scalable and general-purpose agent learning.

Despite their promise, learning directly from real-world websites introduces fundamental obstacles. First, such exploration is often \textbf{unsafe} or \textbf{restricted}: agent actions may interfere with other users, violate platform policies, or be blocked by mechanisms such as Cloudflare and CAPTCHAs. Second, self-generated tasks must be \textbf{well-specified}, \textbf{targeted}, and \textbf{executable}. Poorly specified or ill-defined tasks can misguide learning and invalidate reward signals. 
Prior work often generates underspecified instructions with multiple valid answers and relies on an LLM-as-a-judge to score trajectories~\citep{zhou2025proposer}. However, such LLM-based evaluation can be error-prone, whereas verification-based rewards are typically more reliable and robust~\citep{garcia2025efficient}.
Without reliable task definitions and verifiable outcomes, self-evolving learning becomes unstable and inefficient. Consequently, effective self-evolving web agents critically depend on \textbf{both safe environments and verifiable task construction}.

We introduce \method, a framework that automatically constructs safe, verifiable training environments for self-evolving web agents.
{As in Figure~\ref{fig:motivating}}, rather than training agents directly on real-world websites, \method uses a coding agent to automatically clone a target website into a fully executable synthetic environment, including its frontend, backend logic, and underlying database.
This access allows tasks to be \textbf{generated alongside executable validation programs}~\citep{zhou2025self, wilf2025propose}, enabling automatic validity checks and deterministic evaluation of agent trajectories. As a result, agents trained with \method learn from \textbf{reliable, reproducible training signals} rather than heuristic or LLM-based judgments. By decoupling self-evolving learning from unsafe real-world exploration and grounding it in verifiable environments, \method provides a practical and scalable foundation for training autonomous web agents.

In our experiments, we evaluate \method from two complementary perspectives. First, using WebArena~\citep{zhou2024webarena} and Mind2Web-Online~\citep{xue2025illusion}, we demonstrate that agents trained within our framework generalize to out-of-domain settings and realistic web tasks; on WebArena, \method improves success rates by $+6.06$ (Qwen3-4B) and $+9.09$ (LLaMA-3.2-3B-Instruct) points over the corresponding base models. Second, we investigate whether an agent can achieve site-specific mastery through repeated training within a simulated environment cloned from a fixed website. Beyond these settings, we compare verifiable task generation against prior approaches~\citep{zhou2025proposer}, which generate tasks without direct environment access and rely on LLM-as-a-judge for trajectory evaluation. Our analysis highlights the importance of executable, verifiable tasks for stable agent learning and shows that agent performance improves as the number of training environments increases, indicating the effectiveness of environment scaling in self-evolving web agents.

Our contributions are summarized as follows:
\begin{itemize}
    \item We propose \method, a framework that automatically reconstructs real-world websites into executable synthetic environments and generates verifiable tasks, enabling safe and reliable self-evolving agent learning.
    \item Through extensive experiments on WebArena and Mind2Web-Online, we show that agents trained within \method generalize effectively to unseen websites.
    \item We provide systematic analyses demonstrating the importance of verifiability in task construction and reward assignment, as well as the impact of environment scaling and coding agents on agent learning.
\end{itemize}
\section{Related Work}

\begin{figure*}[!th]
    \centering
    \includegraphics[width=0.9\linewidth]{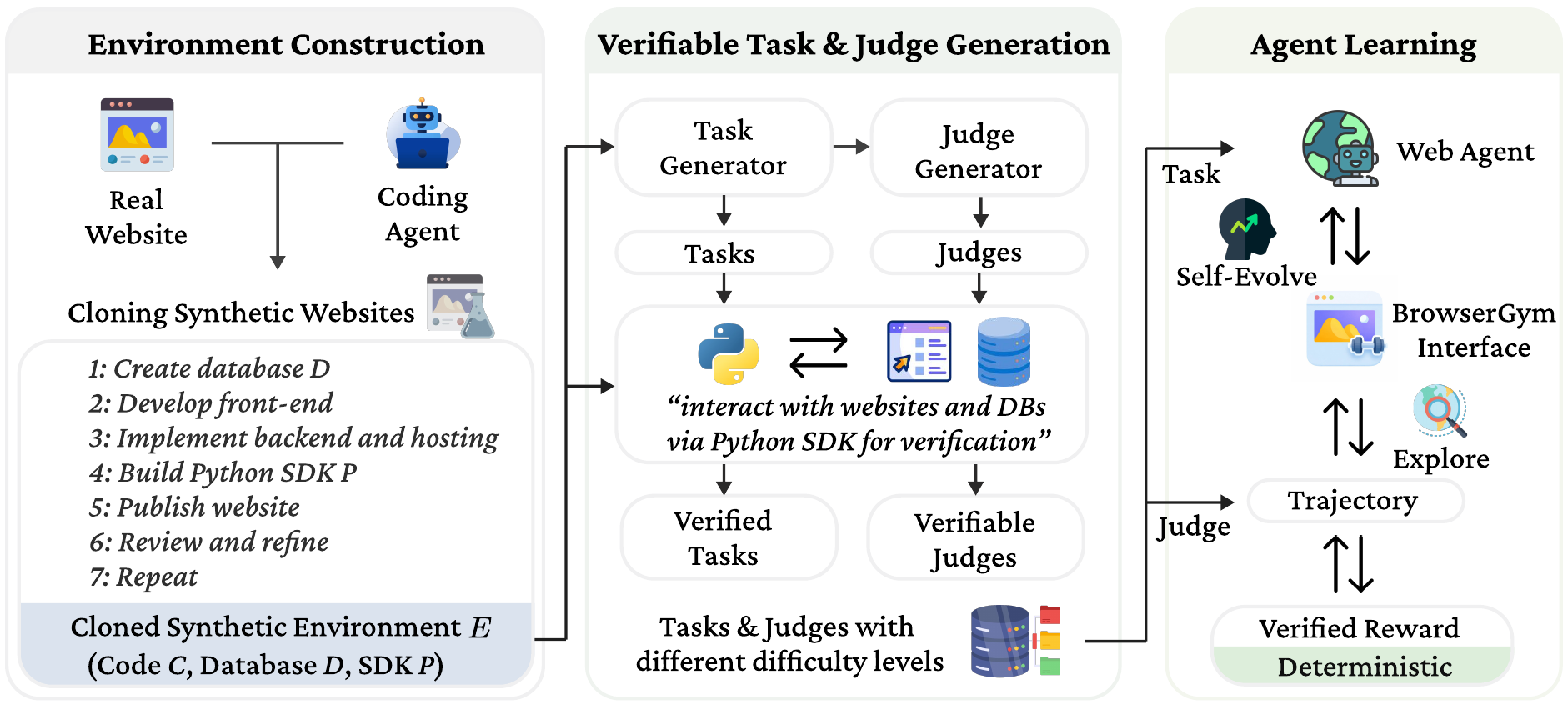}
    \caption{Overview of \method. \method first clones a real website into a fully instrumented synthetic environment (code $C$, database $D$, and a Python SDK $P$) via coding agent, then uses task and judge generators to produce tasks at varying difficulty and verify both tasks and judges by interacting with the website and database through the SDK, yielding deterministic, verified rewards for agent learning.}
    \label{fig:overview}
\end{figure*}

\paragraph{Agent learning with verifiable reward.}
Learning agents for web interaction and tool use typically requires long-horizon trajectories with many sequential decisions, making learning signals sparse and brittle in unconstrained environments.
Recent progress has therefore emphasized \emph{verifiable} training signals and controlled settings where success can be evaluated reliably~\citep{wilf2025propose}.
In math and coding, reinforcement learning with verifiable rewards improves reasoning and tool use by grounding learning in outcome-checkable feedback~\citep{mai2025agent,wen2025reinforcement}.
Beyond single-shot problem solving, self-challenging setups further strengthen supervision by generating executable verifiers and tests~\citep{zhou2025self}.
For web agents, structured pipelines that separate proposing, executing, and evaluating actions offer clearer reward semantics and more scalable skill acquisition~\citep{zhou2025proposer}.
In contrast, \method targets web settings where direct exploration is unsafe or blocked and outcomes are not externally verifiable, by cloning the full website (including its database) and enabling controlled internal validation for trajectory evaluation and reliable rewards.

\paragraph{Self-evolving agents.}
A complementary line of work studies how agents can \emph{self-evolve} via exploration, curricula, and automated task construction, reducing reliance on static human supervision. In realistic benchmarks for web agents such as Mind2Web~\citep{deng2023mind2web}, WebVoyager~\citep{he2024webvoyager}, and WebArena~\citep{zhou2024webarena} enable systematic study of end-to-end agents and iterative improvement. Building on these environments, methods increasingly use online curricula and self-evolving loops: WebRL adapts training tasks to target an agent’s weaknesses over time~\citep{qi2025webrl}, while other work scales coverage via exploration-driven task generation~\citep{ramrakhya2025scaling} or environment/task generation pipelines~\citep{hu2025agentgen}.


Similar self-evolution ideas also appear in reasoning-centric agents: corpus-grounded self-play induces automatic curricula~\citep{liu2025spice}, and reinforced self-training iteratively improves models using self-generated data with reinforcement-style filtering~\citep{gulcehre2023reinforced}. Whereas prior web-agent methods often rely on real-site interaction or unverifiable task generation, \method clones real sites into executable environments with database-backed verification, enabling valid self-generated tasks and fully verifiable rewards without impacting real users or platform constraints.

\paragraph{Coding agents for web development.}

Recent coding agents have demonstrated the ability to autonomously develop web applications end-to-end, ranging from frontend design and backend implementation to deployment~\citep{sweagent, jimenez2024swebench}, by leveraging tool calling for file system access, terminal execution, and external search~\citep{wang2025openhands}. Despite their growing capabilities, such agents frequently introduce implementation errors and require iterative debugging~\citep{chen2024teaching}, which they typically address by incorporating feedback from compiler outputs, runtime logs, language servers, and vision–language models~\citep{muennighoff2023octopack, chae-etal-2024-coffee,zheng2024opencodeinterpreter}.
However, many critical bugs cannot be caught by static checks alone: functional failures, layout issues, and interaction errors often only appear during execution. Prior work therefore, detects such bugs via website interaction using web agents and browser-based testing frameworks~\citep{wang2025openhands,lu2025webgen, lu2025webgenbench}. Building on this, we pair coding agents with automated web interaction to iteratively refine cloned sites, improving functionality and producing reliable synthetic environments.


\section{Method}
Our framework focuses on carefully preparing reliable environments where agents can safely train. We show the overall flow of our framework in Figure~\ref{fig:overview}, where we (i) clone real-world websites into executable synthetic environments~(\cref{subsection:method_cloning}), (ii) derive verifiable tasks and judges from these environments~(\cref{subsection:method_task_generation}), and (iii) train agents on the resulting tasks within the synthetic environments~(\cref{subsection:method_training}).

\subsection{Recreating Real-World Websites}
\label{subsection:method_cloning}
We leverage a coding agent, GPT-5.2~\citep{openai_gpt5.2}, to construct a training environment that ensemble a target website in real-world.
Specifically, given screenshots of a real-world website $E$, a coding agent is tasked with reconstructing the service into a synthetic environment $\tilde{E}$. Toward that goal, the coding agent operates with local file system and terminal access, allowing it to freely write, execute, and iteratively refine code. Through this process, the agent produces an executable system that captures the core application logic and data semantics of the target service.

We represent the resulting synthetic environment $\tilde{E}$ as a tuple $(\mathcal{C}, \mathcal{D}, \mathcal{P})$, where $\mathcal{C}$ denotes the executable application code, $\mathcal{D}$ the underlying database state, and $\mathcal{P}$ a Python SDK that exposes controlled internal access for querying and verifying environment states. In addition to implementing the main application logic, the coding agent also creates auxiliary scripts for environment control, such as bash scripts for server startup and reset utilities, which facilitate repeated experimentation and agent training.

Because the reliability and interface complexity of websites are crucial for training agents, it requires complex programming and debugging process to ensure quality. Thus, after the initial implementation, the cloned environment is further refined through an iterative stabilization process. Imitating human developers' work flow~\citep{lu2025webgen, lu2025webgenbench}, the coding agent is encouraged to interact with the deployed website using Playwright MCP~\citep{playwright_mcp}, identify functional discrepancies, and incrementally patch bugs based on observed failures. This iterative refinement results in a stable and resettable synthetic environment suitable for reliable task execution, validation, and downstream agent learning. While the cloned environment is not perfectly identical to the original website, it preserves the functional structure necessary for verifiable and reproducible training.

\subsection{Verifiable Task and Judge Generation}
\label{subsection:method_task_generation}
Given a synthetic environment $\tilde{E} = (\mathcal{C}, \mathcal{D}, \mathcal{P})$, we prompt large language models (LLMs) to generate tasks that can be automatically verified within the environment. Each task $\mathcal{T}$ is specified by a natural language description and a validation program using the Python SDK $\mathcal{P}$. The goal of this program is to (1) validate the executability of the generated task, and (2) create a verifiable judge. Each task includes a validation program, which specifies task success conditions using executable predicates over environment state. At the end of an episode, these predicates are instantiated as a verifiable judge, which deterministically evaluates the terminal state and returns a binary reward indicating task completion.

\begin{figure}[!t]
    \centering
    \includegraphics[width=1.0\linewidth]{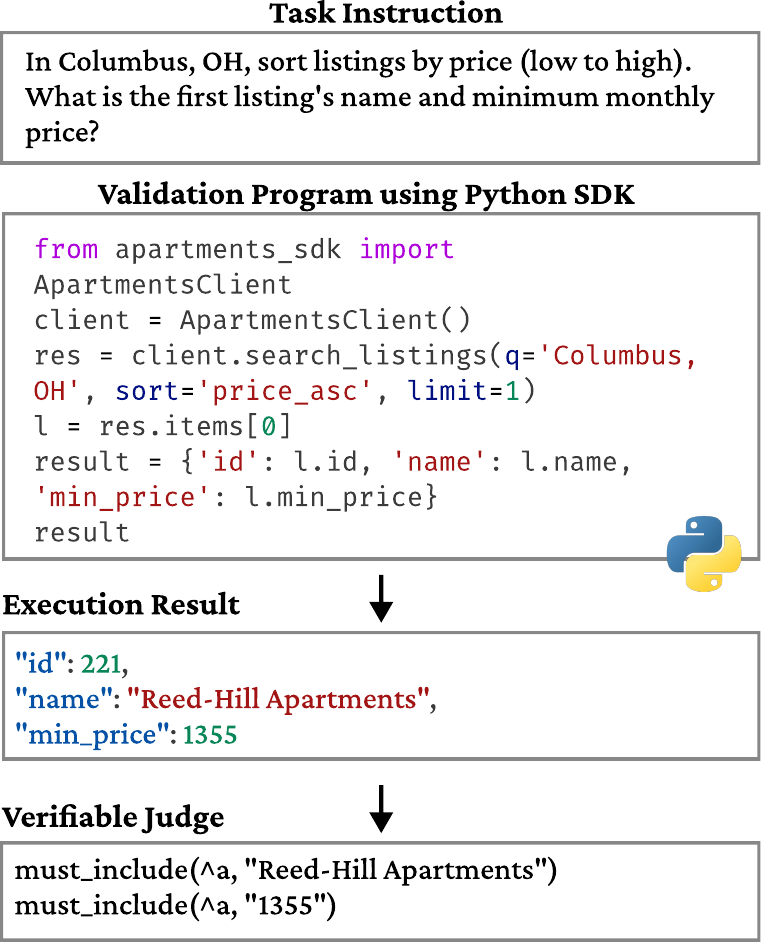}
    \caption{Example of a verifiable task with executable validation in a synthetic recipe website (\ie cloned from apartments.com).}
    \label{fig:task_example}
\end{figure}

For example, in Figure~\ref{fig:task_example}, the task is to sort the list of apartments by price, and answer the name of the first item and its price.
The following validation program first checks whether the task is valid by simulating the desired process, and returns the information to construct the verifiable judge (\eg \texttt{must\_include("Reed-Hill Apartments")}).
This process enables scalable task generation without manual annotation, while guaranteeing that task correctness can be deterministically assessed through executable verification rather than heuristic or LLM-based judgments. Figure~\ref{fig:task_example} provides a concrete example of such a verifiable task, illustrating how natural language instructions are paired with executable validation programs. Such validation programs are subsequently used to compute deterministic reward signals during self-evolving agent learning, as described in the next section.


\subsection{Self-Evolving Agent Learning in Verifiable Environments}
\label{subsection:method_training}
Building on the automatically generated and verifiable tasks, agents are trained through a self-evolving learning loop within the synthetic environment $\tilde{E}$. At each iteration, an agent interacts with the cloned website to solve a sampled task $\mathcal{T}$, producing a trajectory $\tau$ consisting of browser actions and observations.

Upon task completion, the agent’s trajectory $\tau$ is evaluated by executing the task-specific validation program through the Python SDK $\mathcal{P}$, which deterministically queries the underlying database state $\mathcal{D}$. This evaluation yields reproducible reward signals that are independent of heuristic or LLM-based judgments. The verified rewards are then used to update the agent, enabling stable and scalable learning without manual annotations or human supervision. We choose reward-based rejection fine-tuning as an example of a possible training method for utilizing the verifiable rewards.
To support continual self-improvement, newly generated tasks and collected trajectories are iteratively incorporated into the training process. This self-evolving procedure allows agents to progressively adapt to increasingly complex behaviors while remaining grounded in verifiable environment feedback.

\subsection{Environment Statistics and Human Evaluation}
\label{ssec:method_humaneval}
\begin{table}[t]
\caption{Statistics of constructed synthetic environments and generated tasks.}
\label{tab:environment-statistics}
\centering
\small
\begin{tabular}{p{0.60\columnwidth} p{0.30\columnwidth}}
\toprule
\textbf{Statistic} & \textbf{Value} \\
\midrule
Number of websites & 149 \\
Number of tasks per website & 49.5 \\
Total number of tasks & 7,400 \\
Easy tasks & 2,972 (40.2\%) \\
Medium tasks & 	2,900 (39.2\%) \\
Hard tasks & 1,528 (20.6\%) \\
\bottomrule
\end{tabular}
\end{table}
\begin{table*}[t]
\centering
\small
\begin{tabular}{lccccc}
\toprule
Dataset / Benchmark & \# Websites & \# Tasks & Browser Interaction & Verifiable Judge & Scalable Task Gen. \\
\midrule
WebArena & 5 & 812 & \greencheck & \greencheck & \redcross \\
WorkArena & 1 & 33 & \greencheck & \greencheck & \redcross \\
WebVoyager & 15 & 643 & \greencheck & \redcross & \redcross \\
Mind2Web & 137 & 2,350 & \redcross & \greencheck & \redcross \\
Mind2Web-Online & 136 & 300 & \greencheck & \redcross & \redcross \\
Mind2Web-Live & 137 & 542 & \greencheck & \redcross & \redcross \\
\midrule
\textbf{VeriEnv (Ours)} & \textbf{149} & \textbf{7,400} & \greencheck\ (w/ synthetic websites) & \greencheck & \greencheck \\
\bottomrule
\end{tabular}
\caption{Comparison with existing web agent datasets and benchmarks. \method uniquely enables verifiable evaluation and scalable task generation through executable synthetic environments.}
\label{tab:dataset_comparison}
\end{table*}
\begin{table}[!t]
\caption{Human evaluation of the generated websites and tasks.}
\label{tab:human-eval-results}
\centering
\small
\renewcommand{\arraystretch}{1.08}
\begin{tabular}{@{}p{0.68\columnwidth}r@{}}
\toprule
\textbf{Metric} & \textbf{Result} \\
\midrule
\rowcolor{gray!10}\multicolumn{2}{@{}l@{}}{\textbf{Environment quality}} \\
Functional correctness (avg.) & 90\% \\
\hspace{1.5em}Signup & 94\% \\
\hspace{1.5em}Login & 95\% \\
\hspace{1.5em}Search & 81\% \\
\hspace{1.5em}Filter & 88\% \\
\hspace{1.5em}Navigation & 100\% \\
\hspace{1.5em}Forms & 100\% \\
Visual rating (Likert, 1--5) & 4.7 \\
\addlinespace[0.2em]
\midrule
\rowcolor{gray!10}\multicolumn{2}{@{}l@{}}{\textbf{Task validity}} \\
Task executability & 90\% \\
Judge correctness & 76\% \\
\bottomrule
\end{tabular}
\end{table}
We construct synthetic environments for 149 websites, selected by referencing the website list used in Mind2Web~\citep{deng2023mind2web} and Mind2Web-Online~\citep{xue2025illusion} to ensure coverage of realistic and diverse web domains. For each website, we generate 50 task instructions using large language models, resulting in a total of 7,400 tasks. Each task is annotated with a difficulty label (\textbf{easy}, \textbf{medium}, \textbf{hard}) based on predefined criteria reflecting action length, statefulness, and authentication requirements. Table~\ref{tab:environment-statistics} summarizes the statistics of the constructed environments and generated tasks. Also, in Table~\ref{tab:dataset_comparison}, we compare our environments with existing datasets and benchmarks. With our website recreation pipeline, we provide more diverse websites, while the instructions are verifiable.

To conduct a human evaluation, we recruited four graduate students with computer science background. 
Two annotators evaluate 15 instances each in one subset and two evaluate 15 instances each in a second subset, yielding double annotation per subset. 
Annotators rate environment quality via \textbf{Functionality} (success rate over signup, login, search, filter, navigation, and forms) and \textbf{Visual rating} (5-point Likert, higher is better). 
They also assess \textbf{task validity} with binary judgments of whether a task is executable on the synthetic website as described (\textbf{Task executability}) and whether the automated validator correctly determines task completion (\textbf{Judge correctness}). 

Table~\ref{tab:human-eval-results} summarizes the results: functionality averages 90.3\% success across capabilities and visual quality is rated 4.7/5, indicating high-quality synthetic websites. 
Task executability and judge correctness are 90\% and 76\%, respectively. 
The most common errors in judge correctness arise from database resets that do not preserve the random seeds used for populating website data. We find that such errors can be reliably detected and resolved by re-running the validation programs implemented with the Python SDK.
Inter-annotator agreement on the binary judgments is substantial, with mean Cohen’s $\kappa = 0.61$~\citep{cohen1960coefficient}. 
Although judge correctness is lower than task executability, it remains informative because our validators are fully verifiable and rule-based. 

Unlike model-based evaluators (\eg LLM-as-a-Judge~\citep{xue2025illusion}) that can introduce additional uncertainty in complex web environments, these checks yield deterministic, auditable pass/fail decisions when applicable, providing a conservative but reliable foundation for evaluation.



\begin{table*}[!t]
\caption{WebArena-Lite evaluation results across different websites. Methods annotated with * use numbers reported by \citet{chae2025webshepherd}.}
\label{tab:webarena-results}
\centering
\small

\resizebox{0.75\linewidth}{!}{
\begin{tabular}{lccccccc}
\toprule
\textbf{Method} & \textbf{Shopping} & \textbf{CMS} & \textbf{Reddit} & \textbf{GitLab} & \textbf{Map} & \textbf{Total} & $\boldsymbol{\Delta}$ \\
\midrule
GPT-4o-mini~\citep{hurst2024gpt}* 
& 21.74 & 22.86 & 19.05 & 34.38 & 19.35 & 23.64 & -- \\
GPT-4o~\citep{hurst2024gpt}*
& 23.91 &31.43 &28.57 &56.25 &19.35 &31.52&-- \\
\midrule
Qwen3-4B 
& 3.77 & 6.67 & 4.17 & 13.89 & 14.29 & 7.88 & -- \\
\quad +Synatra~\citep{ou2024synatra} 
& 0.00 & 0.00 & 12.50 & 8.33 & 0.00 & 3.64 & $-4.24$ \\
\quad +ADP~\citep{song2025agent} 
& 4.35 & 5.71 & 9.52 & 3.13 & 9.68 & 6.06 & $-1.82$ \\
\quad +\method (Ours) 
& 4.35 & 20.00 & 23.81 & 12.50 & 16.13 & 13.94 & \textbf{+6.06} \\
\midrule
LLaMA-3.2-3B-Instruct 
& 0.00 & 2.86 & 9.52 & 3.13 & 3.23 & 3.03 & -- \\
\quad +Synatra~\citep{ou2024synatra} 
& 2.17	& 2.86& 	14.29&	9.38&	6.45&	6.06& +3.03  \\
\quad +ADP~\citep{song2025agent} 
& 4.35 & 11.43 & 14.29 & 12.50 & 6.45 & 9.09 & +6.06 \\
\quad +\method (Ours) 
& 4.35 & 17.14 & 19.05 & 15.63 & 12.90 & 12.73
& \textbf{+9.70} \\
\bottomrule
\end{tabular}
}
\end{table*}

\begin{table*}[!th]
\caption{Mind2Web-Online results across difficulty levels. Methods annotated with * use numbers reported by \citet{xue2025illusion}.}
\label{tab:mind2web-online-results}
\centering
\small
\resizebox{0.75\linewidth}{!}{
\begin{tabular}{lccccc}
\toprule
\textbf{Method} & \textbf{Easy} & \textbf{Medium} & \textbf{Hard} & \textbf{Total} & $\boldsymbol{\Delta}$ \\
\midrule
Browser-Use-GPT-4o~\citep{browseruse2024}* 
& 55.40 & 26.60 & 8.10 & 30.00 & -- \\
Claude-3.5-Sonnet~\citep{anthropic_claude35_sonnet}* 
& 56.60 & 26.60 & 6.80 & 28.80 & -- \\
\midrule
Qwen3-4B 
& 26.32 & 9.41 & 11.63 & 13.18 & -- \\
\quad +Synatra~\citep{ou2024synatra} 
& 35.09 & 5.88 & 9.30 & 14.55 & $+1.37$ \\
\quad +ADP~\citep{song2025agent} 
& 26.32 & 7.06 & 6.98 & 11.36 & $-1.82$ \\
\quad +\method (Ours) 
& 29.82 & 23.53 & 6.98 & 20.45 & \textbf{+7.27} \\
\midrule
LLaMA-3.2-3B-Instruct 
& 19.30 & 12.94 & 0.00 & 11.36 & -- \\
\quad +Synatra~\citep{ou2024synatra} 
& 24.56	&15.29&	6.98&	14.55& +3.19 \\
\quad +ADP~\citep{song2025agent} 
& 42.11 & 24.71 & 11.63 & 24.09 & +12.73 \\
\quad +\method (Ours) 
& 40.35 & 29.41 & 13.95 & 24.55 & \textbf{+13.19} \\
\bottomrule
\end{tabular}
}
\end{table*}

\section{Experiments}
This section evaluates \method in two complementary settings. First, we study \textbf{cross-domain generalization} by training on recreated websites and testing on established benchmarks that cover unseen sites and tasks. Second, we study \textbf{site-specific mastery} by repeatedly training and self-evolving an agent within a single recreated website to measure in-domain improvements over time.

\subsection{Generalization Across Websites}

\paragraph{Implementation details.}
We implement \method using GPT-5.2~\citep{openai_gpt5.2} as the backbone LLM and Cursor CLI~\citep{cursor_cli} as the coding agent for environment construction. The cloning process takes 83.5 minutes and costs \$3.6 per website on average, including the debugging and task generation process. The list of target websites and the screenshot of the websites are obtained from Mind2Web~\citep{deng2023mind2web}. To evaluate cross-domain generalization, we explicitly exclude websites that overlap with the test split of the evaluation benchmarks from the cloning and training process.

After constructing synthetic environments and generating verifiable tasks, we train web agents based on two open-source base models: Qwen3-4B~\citep{yang2025qwen3} and LLaMA-3.2-3B-Instruct~\citep{dubey2024llama}. To construct training data, we employ a rejection-based fine-tuning strategy on 97 websites. Specifically, for each generated task, we sample agent trajectories and retain only those that successfully satisfy the corresponding executable validation criteria. The resulting filtered trajectories are then used as supervised training data for agent fine-tuning, enabling stable learning from verifiable task completion signals.
Additional implementation details, including training hyperparameters and system configurations, are provided in Appendix~\ref{appendix_ssec:hyperparams}.

\paragraph{Benchmarks and baselines.}
We evaluate agent performance on two widely used benchmarks for web agents: (1) WebArena-Lite~\citep{zhou2024webarena} measures task success across 5 realistic websites implemented within Docker. (2) Mind2Web-Online~\citep{xue2025illusion} focuses on generalization over 100+ real-world websites and provides 300 tasks with three difficulty levels-- easy, medium, and hard. As some of the websites in Mind2Web-Online block web agents, we exclude tasks that have such issues, resulting in 220 tasks. We use the WebJudge-7B model from the original paper for trajectory evaluation.

We consider two categories of baselines. 
(1) \textbf{Proprietary LLMs}: GPT-4o-mini, GPT-4o~\citep{hurst2024gpt} and Claude-3.5-Sonnet~\citep{anthropic_claude35_sonnet}, representing state-of-the-art closed-source models.
(2) \textbf{Open-source web agents}: models trained using existing web-agent datasets and training protocols. In particular, Synatra~\citep{ou2024synatra} constructs synthetic trajectories from website-specific tutorials, while Agent Data Protocol (ADP; \citet{song2025agent}) aggregates multiple agent datasets and standardizes action representations. ADP aggregates diverse web-agent training datasets and provides them in a unified format, simplifying the training process.

\paragraph{Result.}
We show the results in WebArena and Mind2Web-Online on Table~\ref{tab:webarena-results} and Table~\ref{tab:mind2web-online-results}, respectively.
ADP exhibits notably different behaviors depending on the base model. With LLaMA-3.2-3B-Instruct, ADP leads to a clear performance gain, particularly on Mind2Web-Online, which we attribute to dataset overlap between ADP’s {aggregated} training data (including Mind2Web and related web interaction datasets such as NNetNav~\citep{murty2025nnetnavunsupervisedlearningbrowser}) and the evaluation distribution. In contrast, ADP does not consistently benefit Qwen-based models and can even degrade performance. We observe that mixing heterogeneous datasets in ADP introduces issues in generating coherent reasoning and adhering to expected action formats, suggesting a mismatch between ADP’s supervision signals and Qwen’s action-generation behavior.

In contrast, \method consistently improves performance across base models in the fully out-of-domain WebArena setting. We attribute this improvement to training on self-generated trajectories with verified task completion, where successful runs provide structured thought and action tokens as implicit supervision signals. This self-evolving training paradigm encourages more stable learning of reasoning and action token distributions, resulting in improved generalization across both Qwen3-4B and LLaMA-3.2-3B-Instruct, with gains of $+6.06$ and $+9.09$ points, respectively.

\subsection{Site-Specific Mastery via Self-Evolving Training}
\begin{figure*}
    \centering
    \includegraphics[width=0.7\linewidth]{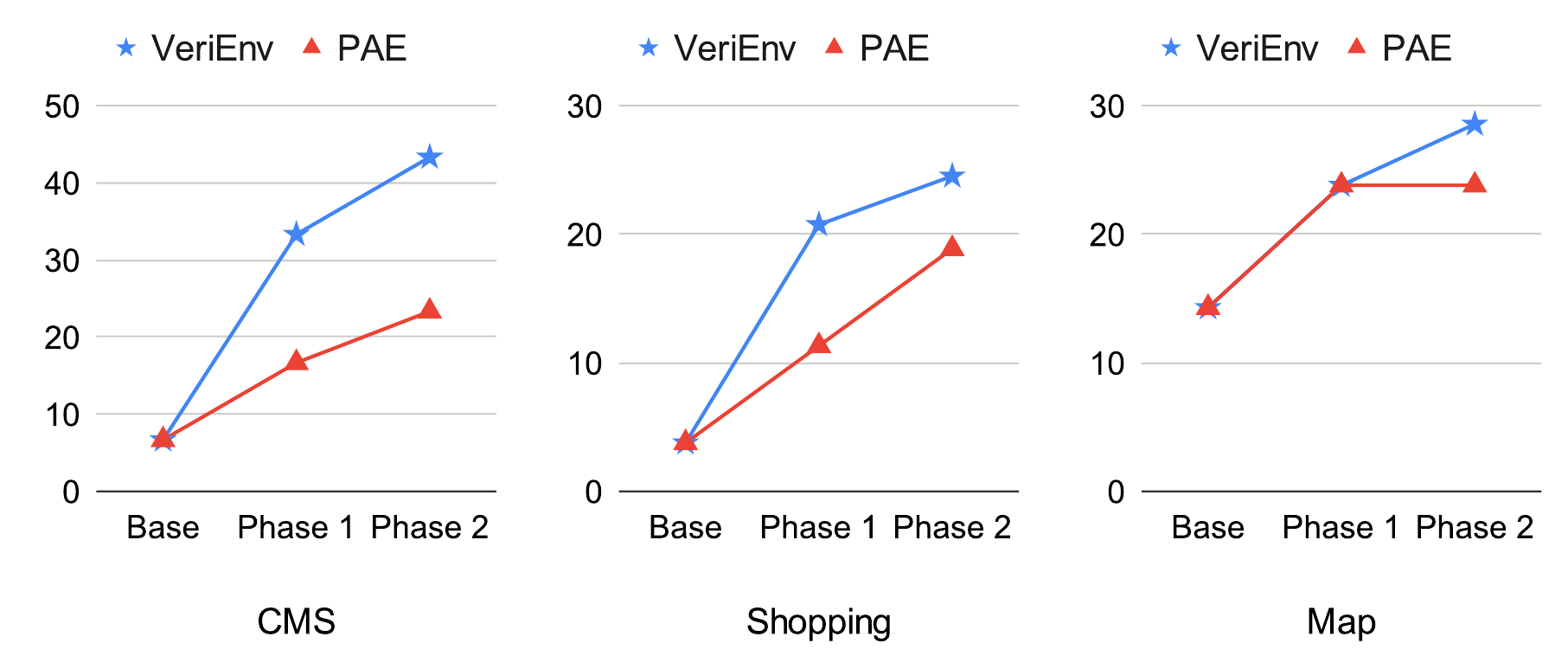}
    \caption{Site-specific self-evolving training within a cloned synthetic environment. 
Agents are trained on a fixed target website using automatically generated tasks and verifiable reward signals.}
    \label{fig:main_indomain}
\end{figure*}
\paragraph{Setup.}
One compelling use case of \method is site-specific mastery, where an agent is trained to excel on a particular website through repeated interaction. In this setting, a target website is cloned into a synthetic environment that serves as an effectively unbounded training gym for self-evolving agents.
To study this scenario, we construct synthetic environments for websites drawn from WebArena and train web agents entirely within the cloned environments. Although WebArena provides sandboxed websites in which agent exploration is inherently safe, we treat these websites as proxies for real-world services. During training, agents are restricted to interact only with the cloned environments rather than the original WebArena instances. The goal of this experiment is to evaluate whether self-evolving training in a verifiable synthetic environment can lead to strong in-domain performance on a fixed website.

We compare \method against PAE~\citep{zhou2025proposer}, a recent approach that generates tasks and uses vision language models for evaluating the trajectories. While both methods leverage automatically generated tasks for agent training, they differ fundamentally in their learning setup. PAE relies on real websites, non-verifiable tasks, and LLM-based judges for reward assignment, whereas \method operates exclusively in synthetic environments and uses verifiable judges to provide deterministic reward signals. 


\paragraph{Result.}
Figure~\ref{fig:main_indomain} presents the results of site-specific self-evolving training across three representative website categories. 
Across all settings, agents trained with \method consistently improve their performance as training progresses from the base model to later self-evolution phases, indicating that repeated interaction within a fixed, cloned environment effectively strengthens in-domain capabilities.
\method yields larger and more stable performance gains than PAE across training phases, with the strongest improvements in CMS and Shopping. While PAE benefits from iterative task generation, its non-verifiable tasks and LLM-judge evaluation constrain progress. \method, by contrast, continues to improve throughout training, consistent with executable, verifiable rewards providing a more reliable learning signal.

These results indicate that verifiable synthetic environments are well-suited for site-specific mastery, enabling agents to progressively refine their behaviors without requiring direct interaction with real-world websites.
Unlike robotics domains, where sim-to-real gaps often pose a fundamental challenge~\citep{Peng2017SimtoRealTO}, web environments exhibit a much smaller discrepancy between synthetic and real executions when the underlying functionality and state transitions are faithfully reproduced.

\begin{figure}[!t]
    \centering
    \includegraphics[width=0.9\linewidth]{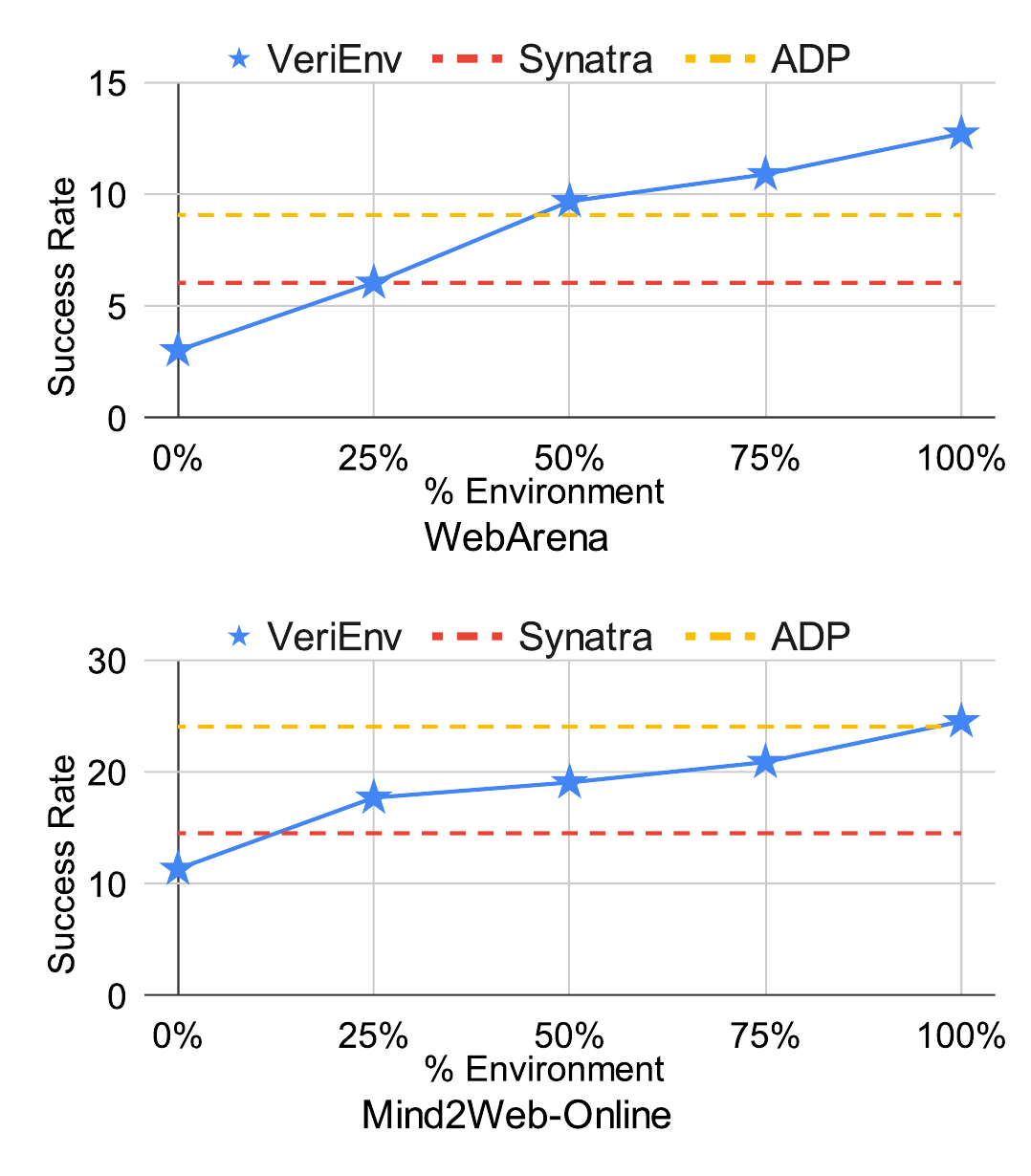}
    \caption{Analysis on the scaling effect of the number of websites.}
    \label{fig:environment_scaling}
\end{figure}
\begin{figure*}[!th]
    \centering
    \includegraphics[width=0.95\linewidth]{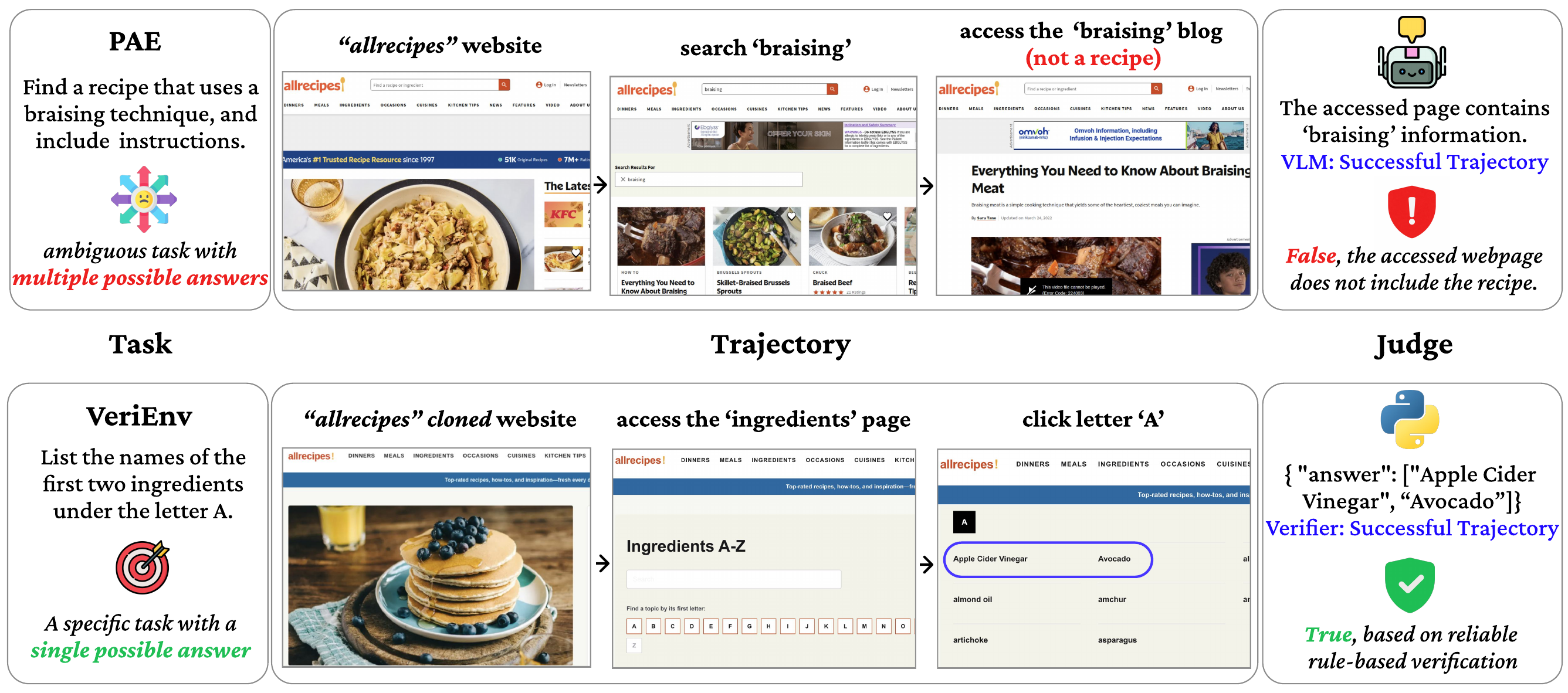}
    \caption{Comparison of task ambiguity and evaluation reliability in PAE~\citep{zhou2025proposer} and \method.}
    \label{fig:example_pae_verienv}
\end{figure*}
\section{Analyses}
This section provides additional analyses to clarify when and why \method works. We study the scaling behavior as the number of recreated websites increases and analyze common failure modes in automated website construction.
\subsection{Environment Scaling Effects on Web Agents}
\method is a fully automatic framework that enables scaling the number of training environments to broaden the coverage of web agents. We analyze how increasing the number of training environments influences agent performance by varying the portion of environments used during training and evaluating intermediate checkpoints on WebArena and Mind2Web-Online.

As shown in Figure~\ref{fig:environment_scaling}, agent performance generally improves as the number of training environments increases across both benchmarks. The improvements follow a consistent upward trend within the evaluated range, indicating that additional environments provide useful learning signals for web agents. In contrast, baseline methods that rely on fixed datasets or non-verifiable supervision exhibit relatively stable performance, suggesting limited sensitivity to environment scaling.
Overall, these results suggest that expanding the diversity of training environments can be beneficial for improving web agent capabilities under verifiable training settings. By enabling safe and systematic environment expansion, \method facilitates a scalable training paradigm that complements existing approaches focused on data or model scaling.

\subsection{Error Analysis on Environment Construction}

\begin{figure}
    \centering
    \includegraphics[width=1.0\columnwidth]{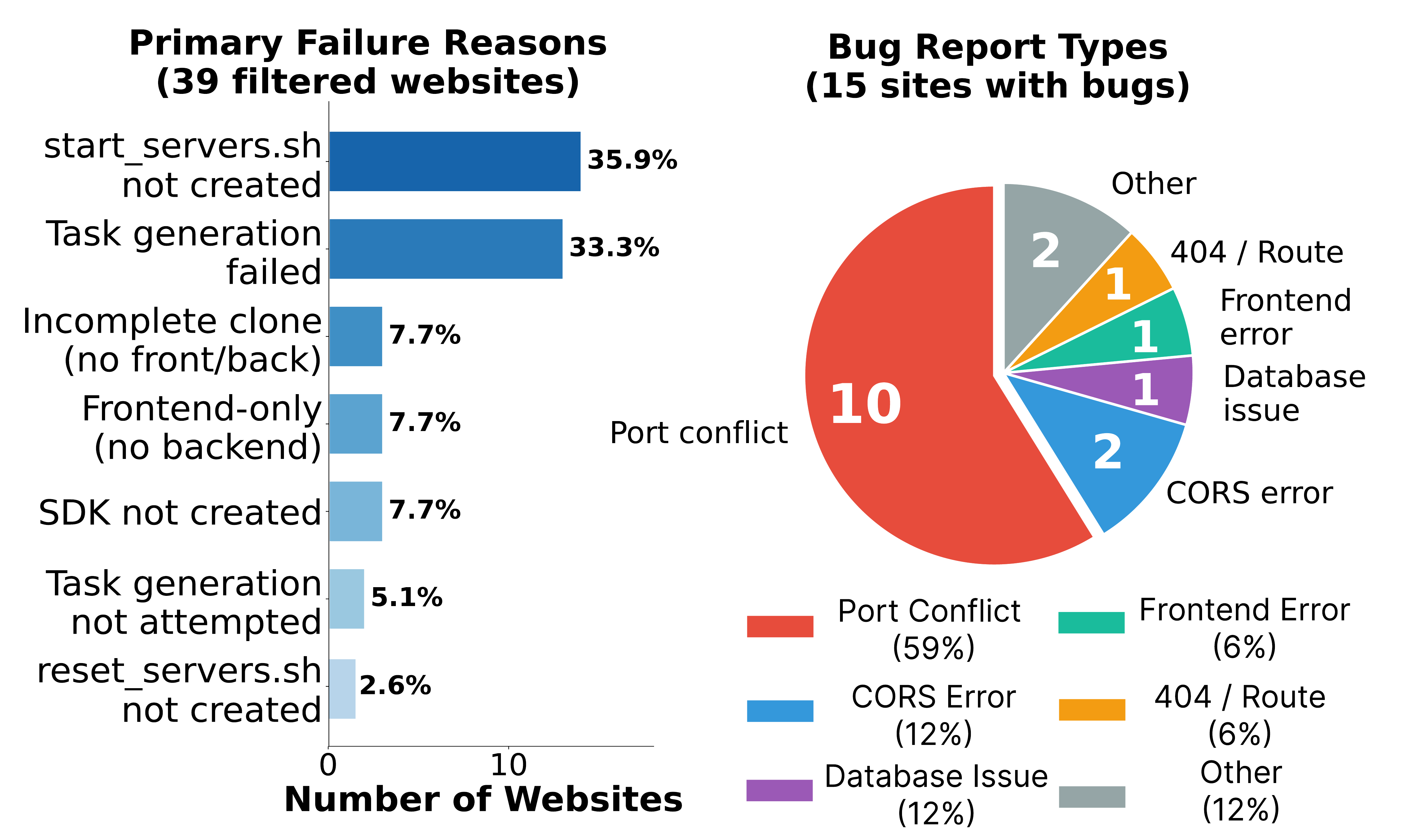}
    \caption{Primary failure reasons for websites excluded and bug types in error report.}
    \label{fig:error_analysis}
\end{figure}

To better understand the limitations of automated environment construction, we analyze the 39 websites (out of 136) that failed to be successfully implemented by our framework. Figure~\ref{fig:error_analysis} summarizes the primary failure modes observed during the implementation and debugging process. The most common issues arise from incomplete system setup, such as missing server startup scripts and failed task generation, indicating that end-to-end orchestration remains a major challenge for coding agents. Among websites that reached a runnable state but still exhibited errors, infrastructure-related issues such as port conflicts and CORS misconfigurations account for the majority of failures. In particular, port conflicts largely stem from deploying more than 100 web applications on a single server, and are not inherent limitations of the approach itself; with sufficient resources, isolating each website by using Docker would be a more reliable and scalable solution.

\subsection{Comparison of \method and PAE}

Figure~\ref{fig:example_pae_verienv} compares PAE~\citep{zhou2025proposer} with \method.
PAE generates tasks from real website interactions and tutorials, but some resulting tasks are ambiguous and admit multiple plausible answers. In such cases, the policy may fail to reach the intended target page (e.g., a recipe), yet a vision-language judge may still label the outcome as successful as long as it contains seemingly relevant information, leading to false positives.
In contrast, \method constructs tasks with a single, well-defined answer and verifies the policy’s terminal state using a rule-based checker backed by a Python SDK, enabling deterministic and reliable evaluation of trajectories.


\section{Discussion and Future Directions}

\subsection{When does a coding agent struggle to recreate websites?}

Although coding agents can recreate a wide range of websites, we observe several recurring scenarios where reconstruction quality degrades. In particular, websites that rely heavily on multimedia delivery are more challenging to reproduce faithfully. Platforms such as arXiv or YouTube involve serving PDF documents or video streams, which require additional infrastructure.

Importantly, these challenges do not fundamentally prevent environment reconstruction. In many cases, the functional behavior of the service can still be approximated by replacing such components with lightweight placeholders. For instance, coding agents can be instructed to serve dummy PDF files or sample video assets, enabling the reconstructed service to remain operational while avoiding the complexity of full media pipelines. Similarly, for image-intensive services such as shopping websites, realistic product catalogs can potentially be generated using modern text-to-image models to populate image databases. This approach could improve the visual realism of synthetic environments without requiring large-scale manual data collection.

\subsection{Future Directions}

An important direction for future work is to leverage these reconstructed environments for reinforcement learning. Because \method provides executable and verifiable judges, the resulting reward signals are deterministic and reproducible, which substantially reduces the instability commonly observed in LLM-based or heuristic evaluation frameworks. We believe this setting enables a more principled study of self-evolving web agents, where agents continuously generate tasks, interact with environments, and improve through scalable training loops.

\section{Conclusion}

We presented \method, which trains web agents in recreated websites by generating tasks with executable, verifiable validators. This design improves safety and reproducibility by avoiding interaction with real services and reducing reliance on subjective LLM judges. Experiments on WebArena and Mind2Web-Online show consistent gains over open-source baselines, and a site-specific setting demonstrates steady improvement through self-evolving training. Overall, our results support environment-centric scaling as a practical route to robust web agents.

\section{Impact Statements}
This work aims to improve the safety, scalability, and reproducibility of web-agent learning by moving data generation and training into recreated websites. By enabling task creation with executable, verifiable validators, the approach can reduce dependence on subjective LLM-based judging and can facilitate more reliable benchmarking and ablation studies.

\textbf{Potential positive impacts.} Recreated websites can support rapid iteration for research on web agents without requiring repeated interaction with real services, which may lower the risk of unintended side effects such as spamming, policy violations, or accidental data modification. The use of deterministic validation can also improve experimental rigor and make agent-training pipelines easier to audit.

\textbf{Risks and negative impacts.} Techniques for recreating websites and training agents in high-fidelity web environments could be misused to develop agents that more effectively automate undesirable behaviors (e.g., large-scale scraping, account abuse, or manipulation of online services). Additionally, recreating websites could raise intellectual-property or terms-of-service concerns if used inappropriately, and recreated environments may inadvertently encode biased or unsafe content present in source sites.

\textbf{Mitigations.} Our framework emphasizes training on recreated environments rather than direct interaction with real services, and it relies on executable validators that can be designed to enforce safety constraints and limit harmful actions during training. To mitigate potential risks, all environments are executed in a sandboxed setting with external network access disabled. The SDK explicitly excludes payment flows, authentication mechanisms, and personally identifiable information, and agents interact solely through simulated browser actions. Internal state exposed via the SDK is used exclusively by the validator for post-hoc evaluation and is never accessible to the agent during execution.

We further encourage responsible use: cloning only websites for which permission is granted (or using internally created templates), limiting the fidelity of sensitive workflows, and releasing models and artifacts with appropriate safeguards (e.g., usage policies, rate limits, and evaluation focused on safety-critical behaviors). Finally, we recommend continued study of transfer from recreated environments to real deployment, including explicit safety evaluations before any real-world use.

\section*{Acknowledgments}
This research is supported in part by the NSF under grant numbers IIS-2052498, SMA-2418946, and NAIRR250217 Any opinions, findings, and conclusions or recommendations expressed in this material are those of the author(s) and do not necessarily reflect the views of the National Science Foundation.

\bibliography{Z_citation}

@misc{anthropic_claude35_sonnet,
  title        = {{Claude 3.5-Sonnet}},
  author       = {{Anthropic}},
  year         = {2025},
  howpublished = {\url{https://www.anthropic.com/news/claude-3-5-sonnet}},
  note         = {Product announcement on the Anthropic website},
}

@misc{browseruse2024,
  title        = {Browser-Use: A Framework for Web Automation with LLMs},
  author       = {{Browser-Use Contributors}},
  year         = {2024},
  howpublished = {\url{https://github.com/browser-use/browser-use}},
  note         = {GitHub repository},
}

@inproceedings{
chae2025webshepherd,
title={Web-Shepherd: Advancing {PRM}s for Reinforcing Web Agents},
author={Hyungjoo Chae and Sunghwan Kim and Junhee Cho and Seungone Kim and Seungjun Moon and Gyeom Hwangbo and Dongha Lim and Minjin Kim and Yeonjun Hwang and Minju Gwak and Dongwook Choi and Minseok Kang and Gwanhoon Im and ByeongUng Cho and Hyojun Kim and Jun Hee Han and Taeyoon Kwon and Minju Kim and Beong-woo Kwak and Dongjin Kang and Jinyoung Yeo},
booktitle={The Thirty-ninth Annual Conference on Neural Information Processing Systems},
year={2025},
url={https://openreview.net/forum?id=G2kMroO9UV}
}

@inproceedings{zheng2024llamafactory,
  title={LlamaFactory: Unified Efficient Fine-Tuning of 100+ Language Models},
  author={Yaowei Zheng and Richong Zhang and Junhao Zhang and Yanhan Ye and Zheyan Luo and Zhangchi Feng and Yongqiang Ma},
  booktitle={Proceedings of the 62nd Annual Meeting of the Association for Computational Linguistics (Volume 3: System Demonstrations)},
  address={Bangkok, Thailand},
  publisher={Association for Computational Linguistics},
  year={2024},
  url={http://arxiv.org/abs/2403.13372}
}

@article{Peng2017SimtoRealTO,
  title={Sim-to-Real Transfer of Robotic Control with Dynamics Randomization},
  author={Xue Bin Peng and Marcin Andrychowicz and Wojciech Zaremba and P. Abbeel},
  journal={2018 IEEE International Conference on Robotics and Automation (ICRA)},
  year={2017},
  pages={1-8},
  url={https://api.semanticscholar.org/CorpusID:3707478}
}

@misc{murty2025nnetnavunsupervisedlearningbrowser,
      title={NNetNav: Unsupervised Learning of Browser Agents Through Environment Interaction in the Wild}, 
      author={Shikhar Murty and Hao Zhu and Dzmitry Bahdanau and Christopher D. Manning},
      year={2025},
      eprint={2410.02907},
      archivePrefix={arXiv},
      primaryClass={cs.CL},
      url={https://arxiv.org/abs/2410.02907}, 
}

@article{hurst2024gpt,
  title={Gpt-4o system card},
  author={Hurst, Aaron and Lerer, Adam and Goucher, Adam P and Perelman, Adam and Ramesh, Aditya and Clark, Aidan and Ostrow, AJ and Welihinda, Akila and Hayes, Alan and Radford, Alec and others},
  journal={arXiv preprint arXiv:2410.21276},
  year={2024}
}

@inproceedings{muennighoff2023octopack,
  title={Octopack: Instruction tuning code large language models},
  author={Muennighoff, Niklas and Liu, Qian and Zebaze, Armel and Zheng, Qinkai and Hui, Binyuan and Zhuo, Terry Yue and Singh, Swayam and Tang, Xiangru and Von Werra, Leandro and Longpre, Shayne},
  booktitle={NeurIPS 2023 workshop on instruction tuning and instruction following}
}

@inproceedings{
chen2024teaching,
title={Teaching Large Language Models to Self-Debug},
author={Xinyun Chen and Maxwell Lin and Nathanael Sch{\"a}rli and Denny Zhou},
booktitle={The Twelfth International Conference on Learning Representations},
year={2024},
url={https://openreview.net/forum?id=KuPixIqPiq}
}

@article{zheng2024opencodeinterpreter,
  title={Opencodeinterpreter: Integrating code generation with execution and refinement},
  author={Zheng, Tianyu and Zhang, Ge and Shen, Tianhao and Liu, Xueling and Lin, Bill Yuchen and Fu, Jie and Chen, Wenhu and Yue, Xiang},
  journal={arXiv preprint arXiv:2402.14658},
  year={2024}
}

@inproceedings{chae-etal-2024-coffee,
    title = "Coffee-Gym: An Environment for Evaluating and Improving Natural Language Feedback on Erroneous Code",
    author = "Chae, Hyungjoo  and
      Kwon, Taeyoon  and
      Moon, Seungjun  and
      Song, Yongho  and
      Kang, Dongjin  and
      Ong, Kai Tzu-iunn  and
      Kwak, Beong-woo  and
      Bae, Seonghyeon  and
      Hwang, Seung-won  and
      Yeo, Jinyoung",
    editor = "Al-Onaizan, Yaser  and
      Bansal, Mohit  and
      Chen, Yun-Nung",
    booktitle = "Proceedings of the 2024 Conference on Empirical Methods in Natural Language Processing",
    month = nov,
    year = "2024",
    address = "Miami, Florida, USA",
    publisher = "Association for Computational Linguistics",
    url = "https://aclanthology.org/2024.emnlp-main.1254/",
    doi = "10.18653/v1/2024.emnlp-main.1254",
    pages = "22503--22524",
}

@inproceedings{sweagent,
 author = {Yang, John and Jimenez, Carlos E. and Wettig, Alexander and Lieret, Kilian and Yao, Shunyu and Narasimhan, Karthik and Press, Ofir},
 booktitle = {Advances in Neural Information Processing Systems},
 doi = {10.52202/079017-1601},
 editor = {A. Globerson and L. Mackey and D. Belgrave and A. Fan and U. Paquet and J. Tomczak and C. Zhang},
 pages = {50528--50652},
 publisher = {Curran Associates, Inc.},
 title = {SWE-agent: Agent-Computer Interfaces Enable Automated Software Engineering},
 volume = {37},
 year = {2024}
}

@inproceedings{
jimenez2024swebench,
title={{SWE}-bench: Can Language Models Resolve Real-world Github Issues?},
author={Carlos E Jimenez and John Yang and Alexander Wettig and Shunyu Yao and Kexin Pei and Ofir Press and Karthik R Narasimhan},
booktitle={The Twelfth International Conference on Learning Representations},
year={2024},
url={https://openreview.net/forum?id=VTF8yNQM66}
}

@article{lu2025webgenbench,
  title={WebGen-Bench: Evaluating LLMs on Generating Interactive and Functional Websites from Scratch},
  author={Lu, Zimu and Yang, Yunqiao and Ren, Houxing and Hou, Haotian and Xiao, Han and Wang, Ke and Shi, Weikang and Zhou, Aojun and Zhan, Mingjie and Li, Hongsheng},
  journal={arXiv preprint arXiv:2505.03733},
  year={2025}
}

@article{lu2025webgen,
  title={Webgen-agent: Enhancing interactive website generation with multi-level feedback and step-level reinforcement learning},
  author={Lu, Zimu and Ren, Houxing and Yang, Yunqiao and Wang, Ke and Zong, Zhuofan and Pan, Junting and Zhan, Mingjie and Li, Hongsheng},
  journal={arXiv preprint arXiv:2509.22644},
  year={2025}
}

@article{huang2025r,
  title={R-zero: Self-evolving reasoning llm from zero data},
  author={Huang, Chengsong and Yu, Wenhao and Wang, Xiaoyang and Zhang, Hongming and Li, Zongxia and Li, Ruosen and Huang, Jiaxin and Mi, Haitao and Yu, Dong},
  journal={arXiv preprint arXiv:2508.05004},
  year={2025}
}

@article{xu2024theagentcompany,
  title={Theagentcompany: benchmarking llm agents on consequential real world tasks},
  author={Xu, Frank F and Song, Yufan and Li, Boxuan and Tang, Yuxuan and Jain, Kritanjali and Bao, Mengxue and Wang, Zora Z and Zhou, Xuhui and Guo, Zhitong and Cao, Murong and others},
  journal={arXiv preprint arXiv:2412.14161},
  year={2024}
}

@article{wilf2025propose,
  title={Propose, Solve, Verify: Self-Play Through Formal Verification},
  author={Wilf, Alex and Aggarwal, Pranjal and Parno, Bryan and Fried, Daniel and Morency, Louis-Philippe and Liang, Paul Pu and Welleck, Sean},
  journal={arXiv preprint arXiv:2512.18160},
  year={2025}
}

@misc{playwright_mcp,
  title        = {{Playwright MCP}},
  author       = {{Microsoft}},
  year         = {2024},
  howpublished = {\url{https://github.com/microsoft/playwright-mcp}},
  note         = {Model Context Protocol integration for Playwright-based browser automation},
}

@misc{cursor_cli,
  title        = {{Cursor CLI Overview}},
  author       = {{Cursor}},
  howpublished = {\url{https://cursor.com/docs/cli/overview}},
  note         = {Official documentation for the Cursor command-line interface (accessed December 2025)},
  year         = {2025},
}

@article{xue2025illusion,
  title={An illusion of progress? assessing the current state of web agents},
  author={Xue, Tianci and Qi, Weijian and Shi, Tianneng and Song, Chan Hee and Gou, Boyu and Song, Dawn and Sun, Huan and Su, Yu},
  journal={arXiv preprint arXiv:2504.01382},
  year={2025}
}

@inproceedings{osworld,
  author={Tianbao Xie and Danyang Zhang and Jixuan Chen and Xiaochuan Li and Siheng Zhao and Ruisheng Cao and Toh Jing Hua and Zhoujun Cheng and Dongchan Shin and Fangyu Lei and Yitao Liu and Yiheng Xu and Shuyan Zhou and Silvio Savarese and Caiming Xiong and Victor Zhong and Tao Yu},
  title={OSWorld: Benchmarking Multimodal Agents for Open-Ended Tasks in Real Computer Environments},
  year={2024},
  cdate={1704067200000},
  booktitle={NeurIPS},
}

@misc{openai_gpt5.2,
  title        = {GPT-5.2},
  author       = {{OpenAI}},
  year         = {2025},
  howpublished = {\url{https://openai.com/index/introducing-gpt-5-2/}},
  note         = {Official announcement of GPT-5.2, OpenAI’s latest large language model with improved reasoning, tool use, and long-context understanding},
}

@inproceedings{
ou2024synatra,
title={Synatra: Turning Indirect Knowledge into Direct Demonstrations for Digital Agents at Scale},
author={Tianyue Ou and Frank F. Xu and Aman Madaan and Jiarui Liu and Robert Lo and Abishek Sridhar and Sudipta Sengupta and Dan Roth and Graham Neubig and Shuyan Zhou},
booktitle={The Thirty-eighth Annual Conference on Neural Information Processing Systems},
year={2024},
url={https://openreview.net/forum?id=KjNEzWRIqn}
}

@article{song2025agent,
  title={Agent data protocol: Unifying datasets for diverse, effective fine-tuning of llm agents},
  author={Song, Yueqi and Ramaneti, Ketan and Sheikh, Zaid and Chen, Ziru and Gou, Boyu and Xie, Tianbao and Xu, Yiheng and Zhang, Danyang and Gandhi, Apurva and Yang, Fan and others},
  journal={arXiv preprint arXiv:2510.24702},
  year={2025}
}

@article{ramrakhya2025scaling,
  title={Scaling synthetic task generation for agents via exploration},
  author={Ramrakhya, Ram and Szot, Andrew and Attia, Omar and Yang, Yuhao and Nguyen, Anh and Mazoure, Bogdan and Gan, Zhe and Agrawal, Harsh and Toshev, Alexander},
  journal={arXiv preprint arXiv:2509.25047},
  year={2025}
}

@inproceedings{qi2025webrl,
  title={WebRL: Training LLM Web Agents via Self-Evolving Online Curriculum Reinforcement Learning},
  author={Qi, Zehan and Liu, Xiao and Iong, Iat Long and Lai, Hanyu and Sun, Xueqiao and Sun, Jiadai and Yang, Xinyue and Yang, Yu and Yao, Shuntian and Xu, Wei and others},
  booktitle={ICLR},
  year={2025}
}

@inproceedings{hu2025agentgen,
  title={Agentgen: Enhancing planning abilities for large language model based agent via environment and task generation},
  author={Hu, Mengkang and Zhao, Pu and Xu, Can and Sun, Qingfeng and Lou, Jian-Guang and Lin, Qingwei and Luo, Ping and Rajmohan, Saravan},
  booktitle={Proceedings of the 31st ACM SIGKDD Conference on Knowledge Discovery and Data Mining V. 1},
  pages={496--507},
  year={2025}
}

@inproceedings{he2024webvoyager,
  title={WebVoyager: Building an End-to-End Web Agent with Large Multimodal Models},
  author={He, Hongliang and Yao, Wenlin and Ma, Kaixin and Yu, Wenhao and Dai, Yong and Zhang, Hongming and Lan, Zhenzhong and Yu, Dong},
  booktitle={Proceedings of the 62nd Annual Meeting of the Association for Computational Linguistics (Volume 1: Long Papers)},
  pages={6864--6890},
  year={2024}
}

@article{deng2023mind2web,
  title={Mind2web: Towards a generalist agent for the web},
  author={Deng, Xiang and Gu, Yu and Zheng, Boyuan and Chen, Shijie and Stevens, Sam and Wang, Boshi and Sun, Huan and Su, Yu},
  journal={Advances in Neural Information Processing Systems},
  volume={36},
  pages={28091--28114},
  year={2023}
}

@inproceedings{zhou2024webarena,
  title={WEBARENA: A REALISTIC WEB ENVIRONMENT FOR BUILDING AUTONOMOUS AGENTS},
  author={Zhou, Shuyan and Xu, Frank F and Zhu, Hao and Zhou, Xuhui and Lo, Robert and Sridhar, Abishek and Cheng, Xianyi and Ou, Tianyue and Bisk, Yonatan and Fried, Daniel and others},
  booktitle={12th International Conference on Learning Representations, ICLR 2024},
  year={2024}
}

@inproceedings{
wang2025openhands,
title={OpenHands: An Open Platform for {AI} Software Developers as Generalist Agents},
author={Xingyao Wang and Boxuan Li and Yufan Song and Frank F. Xu and Xiangru Tang and Mingchen Zhuge and Jiayi Pan and Yueqi Song and Bowen Li and Jaskirat Singh and Hoang H. Tran and Fuqiang Li and Ren Ma and Mingzhang Zheng and Bill Qian and Yanjun Shao and Niklas Muennighoff and Yizhe Zhang and Binyuan Hui and Junyang Lin and Robert Brennan and Hao Peng and Heng Ji and Graham Neubig},
booktitle={The Thirteenth International Conference on Learning Representations},
year={2025},
url={https://openreview.net/forum?id=OJd3ayDDoF}
}

@article{liu2025spice,
  title={Spice: Self-play in corpus environments improves reasoning},
  author={Liu, Bo and Jin, Chuanyang and Kim, Seungone and Yuan, Weizhe and Zhao, Wenting and Kulikov, Ilia and Li, Xian and Sukhbaatar, Sainbayar and Lanchantin, Jack and Weston, Jason},
  journal={arXiv preprint arXiv:2510.24684},
  year={2025}
}

@article{gulcehre2023reinforced,
  title={Reinforced self-training (rest) for language modeling},
  author={Gulcehre, Caglar and Paine, Tom Le and Srinivasan, Srivatsan and Konyushkova, Ksenia and Weerts, Lotte and Sharma, Abhishek and Siddhant, Aditya and Ahern, Alex and Wang, Miaosen and Gu, Chenjie and others},
  journal={arXiv preprint arXiv:2308.08998},
  year={2023}
}

@article{mai2025agent,
  title={Agent rl scaling law: Agent rl with spontaneous code execution for mathematical problem solving},
  author={Mai, Xinji and Xu, Haotian and Li, Zhong-Zhi and Wang, Weinong and Hu, Jian and Zhang, Yingying and Zhang, Wenqiang and others},
  journal={arXiv preprint arXiv:2505.07773},
  year={2025}
}

@article{wen2025reinforcement,
  title={Reinforcement learning with verifiable rewards implicitly incentivizes correct reasoning in base llms},
  author={Wen, Xumeng and Liu, Zihan and Zheng, Shun and Ye, Shengyu and Wu, Zhirong and Wang, Yang and Xu, Zhijian and Liang, Xiao and Li, Junjie and Miao, Ziming and others},
  journal={arXiv preprint arXiv:2506.14245},
  year={2025}
}

@article{zhou2025self,
  title={Self-challenging language model agents},
  author={Zhou, Yifei and Levine, Sergey and Weston, Jason and Li, Xian and Sukhbaatar, Sainbayar},
  journal={arXiv preprint arXiv:2506.01716},
  year={2025}
}

@inproceedings{zhou2025proposer,
  title={Proposer-agent-evaluator (pae): Autonomous skill discovery for foundation model internet agents},
  author={Zhou, Yifei and Yang, Qianlan and Lin, Kaixiang and Bai, Min and Zhou, Xiong and Wang, Yu-Xiong and Levine, Sergey and Li, Li Erran},
  booktitle={Forty-second International Conference on Machine Learning},
  year={2025}
}

@article{cohen1960coefficient,
  title={A coefficient of agreement for nominal scales},
  author={Cohen, Jacob},
  journal={Educational and Psychological Measurement},
  volume={20},
  number={1},
  pages={37--46},
  year={1960},
  doi={10.1177/001316446002000104}
}

@article{yang2025qwen3,
  title={Qwen3 technical report},
  author={Yang, An and Li, Anfeng and Yang, Baosong and Zhang, Beichen and Hui, Binyuan and Zheng, Bo and Yu, Bowen and Gao, Chang and Huang, Chengen and Lv, Chenxu and others},
  journal={arXiv preprint arXiv:2505.09388},
  year={2025}
}

@article{dubey2024llama,
  title={The llama 3 herd of models},
  author={Dubey, Abhimanyu and Jauhri, Abhinav and Pandey, Abhinav and Kadian, Abhishek and Al-Dahle, Ahmad and Letman, Aiesha and Mathur, Akhil and Schelten, Alan and Yang, Amy and Fan, Angela and others},
  journal={arXiv e-prints},
  pages={arXiv--2407},
  year={2024}
}

@inproceedings{garcia2025efficient,
  title={Efficient safety retrofitting against jailbreaking for llms},
  author={Garcia-Gasulla, Dario and Tormos, Adri{\'a}n and Arias-Duart, Anna and Hinjos, Daniel and Molina-Sedano, Oscar and Gurarajan, Ashwin Kumar and Cardello, Maria Eugenia},
  booktitle={International Conference on Computer Safety, Reliability, and Security},
  pages={537--565},
  year={2025},
  organization={Springer}
}
\bibliographystyle{icml2026}

\newpage
\onecolumn
\appendix
\section{Implementation Details of \method}
\subsection{Agent Architecture and Training Hyperparameters}

\subsubsection{Coding Agent and LLMs for Implementing \method}
\label{appendix_ssec:coding_agent}
We experimented with several coding agent systems for implementing \method, including Cursor CLI, Claude CLI, and OpenHands. In practice, we found that both Claude CLI and OpenHands frequently terminated the implementation process prematurely, even when the target website was not fully functional or when critical components such as the Python SDK were missing. These early exits made it difficult to reliably construct complete, production-ready synthetic environments.

We also evaluated alternative backbone language models for environment construction. In addition to GPT-5.2, which we use throughout our experiments, we tested open-source code-oriented LLMs such as Qwen3-Coder-30B-A3B-Instruct and GLM-4.7-Flash. However, we observed that the lack of strong multimodal capabilities significantly limited their effectiveness. In particular, these models struggled to diagnose and fix frontend issues (e.g., layout inconsistencies) and often failed to properly utilize available tools, such as executing shell commands or interacting with websites via Playwright MCP. As a result, they were less reliable for end-to-end website reconstruction in our setting.
\subsubsection{Training Details and Hyperparameters}
\label{appendix_ssec:hyperparams}
We train all web agents using LLaMA-Factory~\citep{zheng2024llamafactory}.
For all experiments, we use a learning rate of $1\times10^{-5}$ and train for two epochs.
We adopt a linear learning rate warmup over the first 10\% of the total training steps.

Training is performed with a maximum sequence length of 8{,}000 tokens.
We use DeepSpeed ZeRO-3 for memory-efficient distributed training, with a gradient accumulation step of 2.
All experiments are conducted using two NVIDIA A40 GPUs.
\begin{figure*}[!h]
    \centering
    \includegraphics[width=0.7\linewidth]{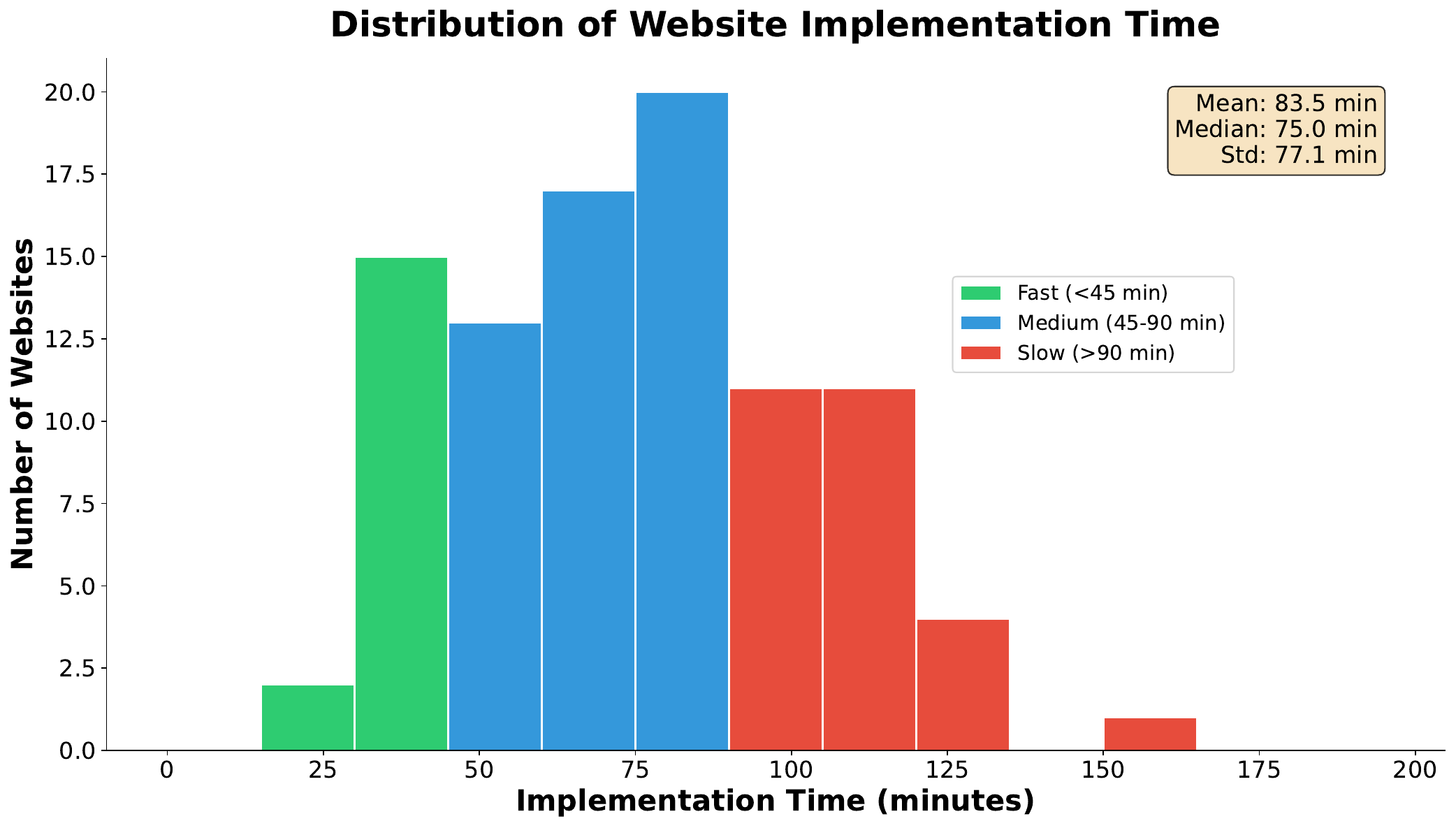}
    \caption{Distribution of website implementation time for constructing synthetic environments using a coding agent. Each bar shows the number of websites grouped by implementation duration, categorized as fast (<45 minutes), medium (45–90 minutes), and slow (>90 minutes). The distribution indicates that most websites can be reconstructed within a moderate time budget, with a long tail corresponding to more complex implementations.}
    \label{fig:implementation_time}
\end{figure*}

\subsection{Synthetic Environment Construction Pipeline}

We provide the prompt used for website reconstruction from snapshots in Figure~\ref{fig:prompt_website_reconstruction}.

We provide the prompt used by the coding agent for website implementation, bug reporting, and debugging in Figure~\ref{fig:prompt_website_implementation_bug_reporting}.

\subsection{Task Generation and Validation Implementation}

We provide the prompt used to generate tasks and validation judges in Figure~\ref{fig:prompt_task_judge}.

\paragraph{Example of Implementation and Debugging Process.}

To illustrate the end-to-end implementation and debugging workflow enabled by \method, we present a concrete example based on cloning a real-world retail website. Starting from a set of reference screenshots, the coding agent first constructs an initial executable version of the website, including frontend pages, backend APIs, database schemas, and a Python SDK that exposes internal functionalities. The initial implementation prioritizes functional completeness, ensuring that all major pages, navigation flows, and APIs are runnable via the provided server management scripts (e.g., \texttt{start\_servers.sh}).

\begin{figure*}
    \centering
    \fbox{\includegraphics[width=0.8\linewidth]{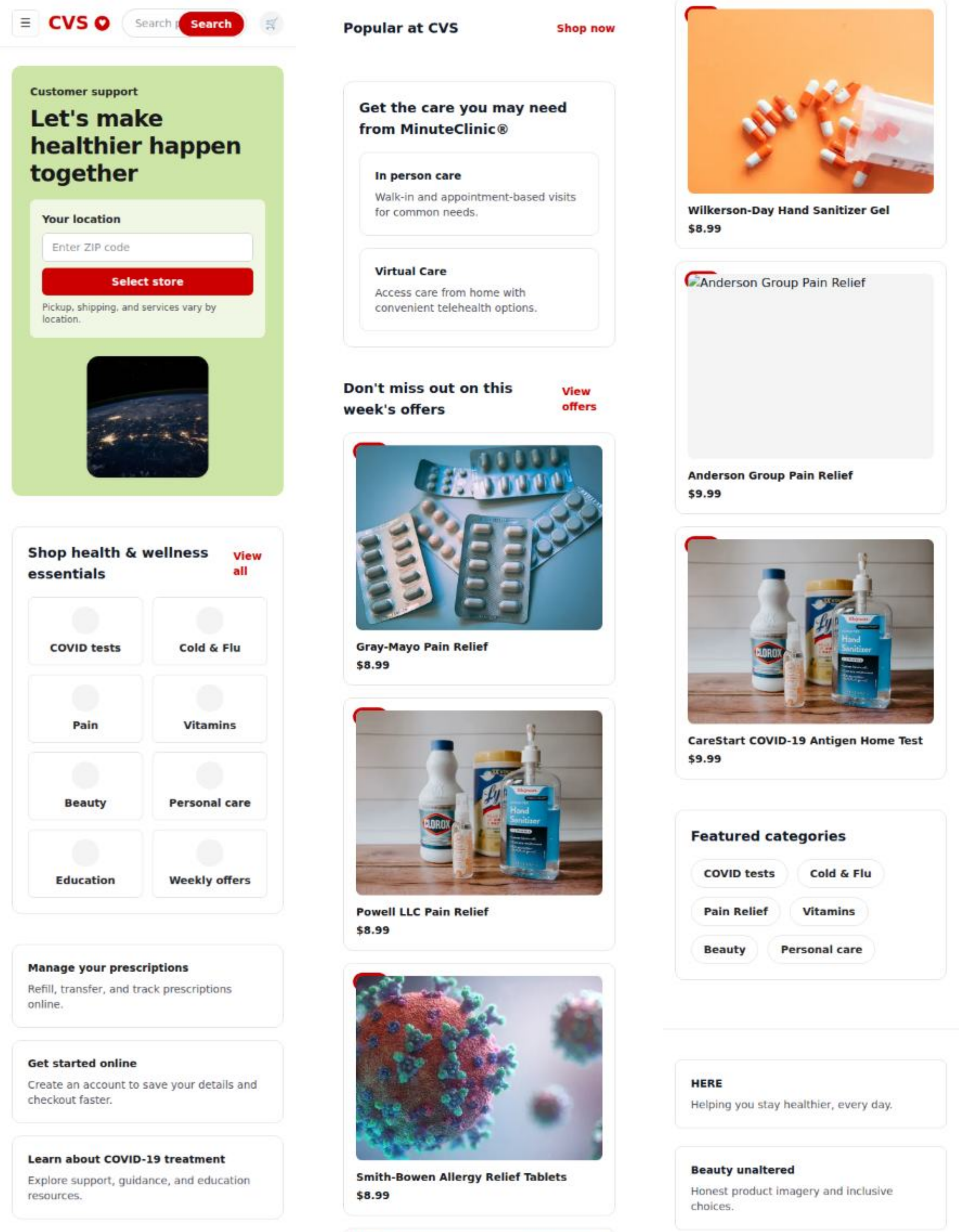}}
    \caption{A browser snapshot attached to the bug report `bug\_report\_2026-01-07\_23-36-45'.}
    \label{fig:bug_report_snapshot}
\end{figure*}

\begin{figure*}
    \centering
    \fbox{\includegraphics[width=0.85\linewidth]{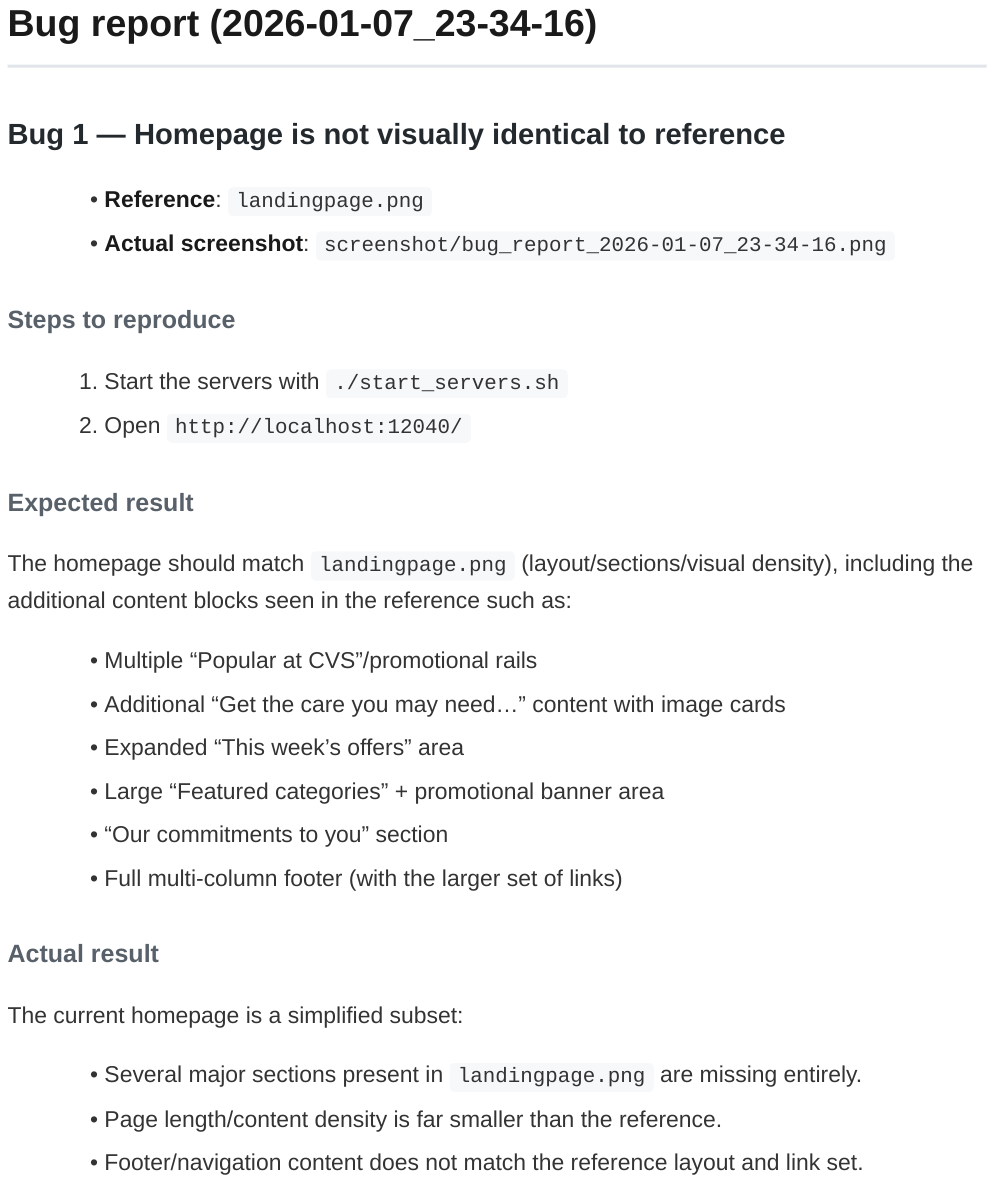}}
    \caption{Bug report example highlighting a missing homepage section discovered during automated traversal.}
    \label{fig:bug_report_homepage}
\end{figure*}

\begin{figure*}
    \centering
    \fbox{\includegraphics[width=0.85\linewidth]{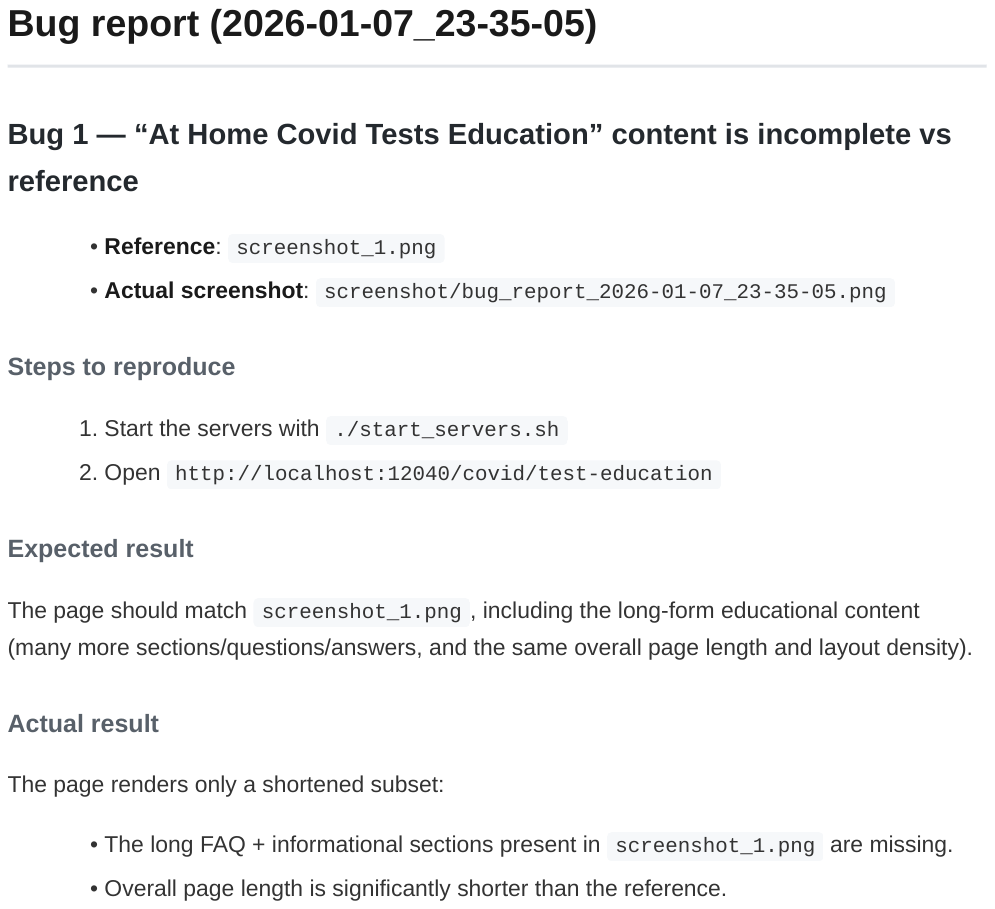}}
    \caption{Bug report example showing incomplete long-form content on an informational page relative to the reference.}
    \label{fig:bug_report_incomplete}
\end{figure*}

\begin{figure*}
    \centering
    \fbox{\includegraphics[width=0.85\linewidth]{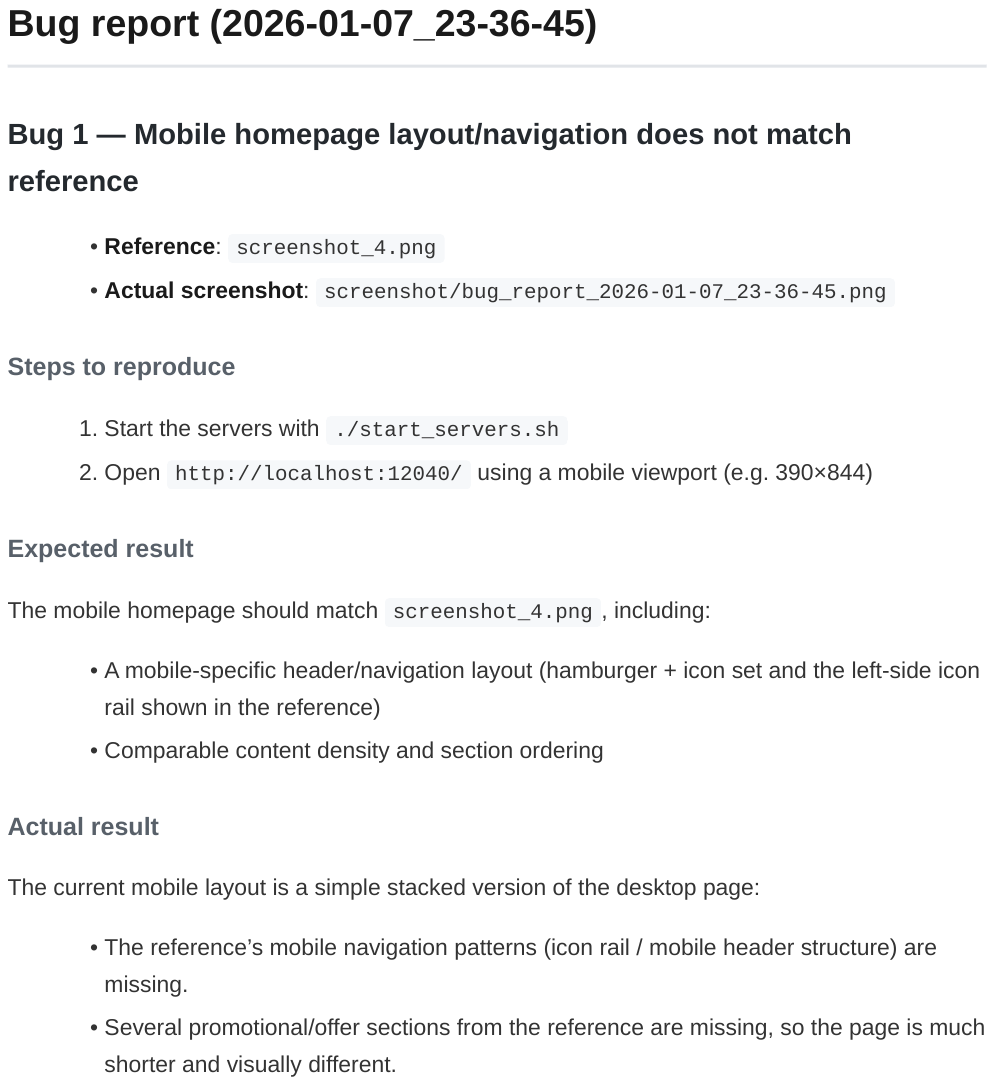}}
    \caption{Bug report example illustrating a desktop--mobile layout mismatch detected across viewports.}
    \label{fig:bug_report_layout}
\end{figure*}

After the initial implementation, the coding agent performs iterative bug discovery using Playwright MCP. The agent systematically explores the synthesized website across different pages and viewports, comparing rendered content against the reference screenshots as in Figure~\ref{fig:bug_report_snapshot}. Discrepancies are documented as structured bug reports that include reproduction steps, expected versus actual behavior, and visual evidence captured as screenshots. The agent identifies issues such as missing homepage sections~(Figure \ref{fig:bug_report_homepage}), incomplete long-form content on informational pages~(Figure \ref{fig:bug_report_incomplete}), and layout inconsistencies between desktop and mobile views~(Figure \ref{fig:bug_report_layout}).

\begin{figure*}
    \centering
    \fbox{\includegraphics[width=0.85\linewidth]{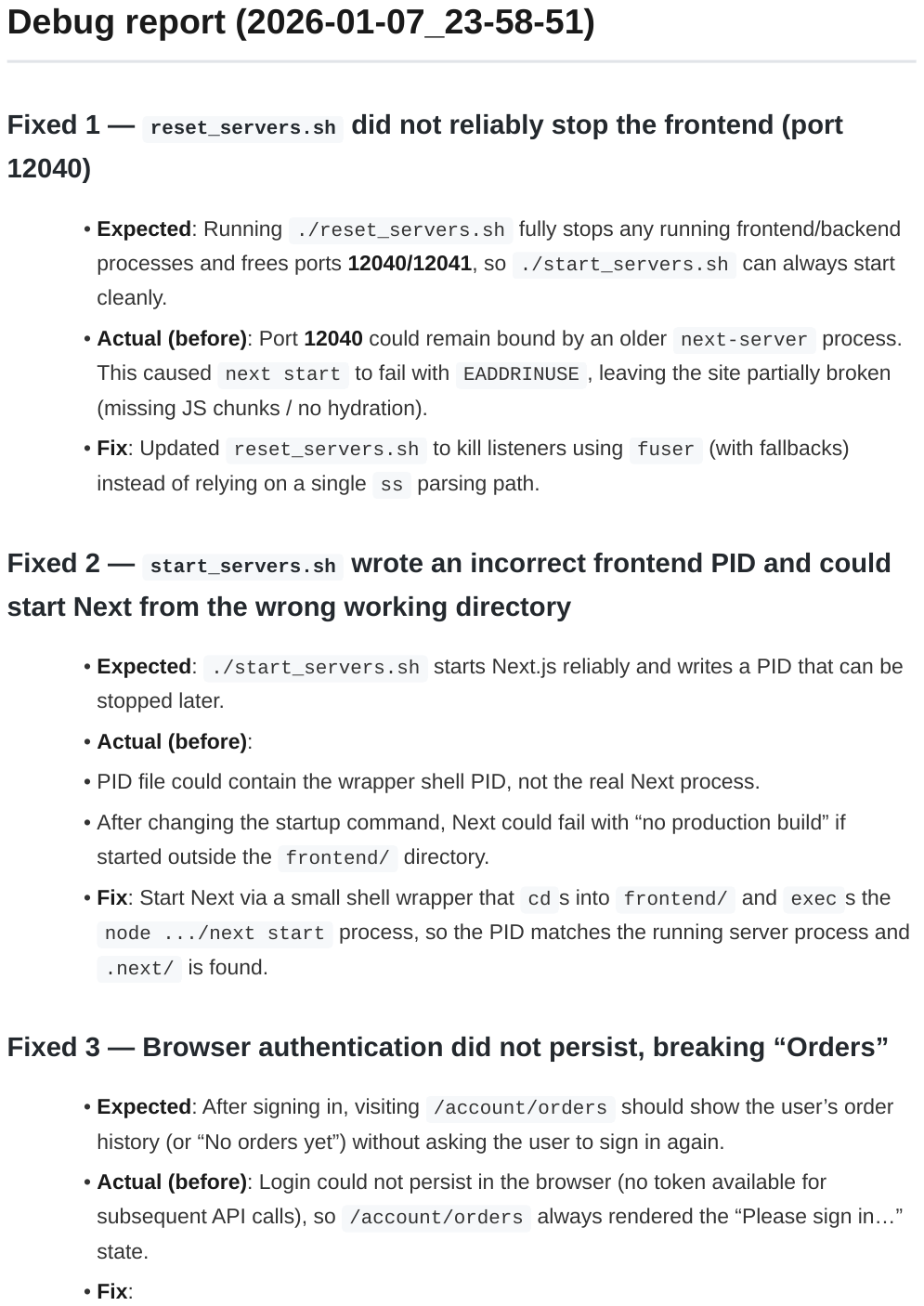}}
    \caption{Structured debug report after patching, summarizing fixes and verification results.}
    \label{fig:debug_note}
\end{figure*}

Following bug reporting, the agent enters a debugging and patching phase. Reported issues are addressed by modifying frontend components, backend logic, and server scripts as needed. For instance, server lifecycle scripts are refined to ensure reliable resets between runs, and authentication-related bugs are fixed to correctly maintain session state across API calls. Visual and content-level refinements are applied to better align the synthesized website with the reference design. Each fix is verified through repeated Playwright-based testing, and the debugging outcomes are recorded in structured debug reports with post-fix screenshots as in Figure~\ref{fig:debug_note}.

This example demonstrates how \method supports a fully automated yet auditable environment construction pipeline, where implementation, bug discovery, and debugging are tightly coupled through executable scripts, visual inspection, and reproducible reports. Such iterative refinement is essential for producing high-fidelity synthetic environments that support verifiable task generation and reliable self-evolving agent learning.

\subsection{Cloned Synthetic Website Examples}

We showcase examples of our cloned synthetic websites. Specifically, we randomly sample four sites
(\eg, CarMax, CVS, Eventbrite, and Google Finance) and present representative screenshots from each
site (Parts 1--3) illustrating key interface elements and interactions.
We map each website’s examples to the corresponding pages in Table~\ref{tab:synthetic-website-fig-links}.

\begin{table}[h]
\centering
\caption{Cloned synthetic website screenshots. Each entry links to the corresponding figure for Parts 1--3.}
\begin{tabular}{lcccc}
\hline
Website & Original URL & Part 1 & Part 2 & Part 3 \\
\hline
CarMax  & \url{https://www.carmax.com/}   & Fig.~\ref{fig:synth-web-1-1} & Fig.~\ref{fig:synth-web-1-2} & Fig.~\ref{fig:synth-web-1-3} \\
CVS     & \url{https://www.cvs.com/}        & Fig.~\ref{fig:synth-web-2-1} & Fig.~\ref{fig:synth-web-2-2} & Fig.~\ref{fig:synth-web-2-3} \\
Eventbrite  & \url{https://www.eventbrite.com/}   & Fig.~\ref{fig:synth-web-3-1} & Fig.~\ref{fig:synth-web-3-2} & Fig.~\ref{fig:synth-web-3-3} \\
Google Finance & \url{https://www.google.com/finance/} & Fig.~\ref{fig:synth-web-4-1} & Fig.~\ref{fig:synth-web-4-2} & Fig.~\ref{fig:synth-web-4-3} \\
\hline
\end{tabular}
\label{tab:synthetic-website-fig-links}
\end{table}


\section{Synthetic Website Evaluation Interface}
\label{appendix:synthetic_website_human_eval}

This section documents the Label Studio\footnote{https://labelstud.io/} annotation interface used to assess (A) the quality of synthetic websites and (B) the validity of generated tasks and their associated validation programs (``judges''). 
Each annotation item provides the annotator with a website URL, a task instruction, and the judge code. 
Annotators interact with the website, record observed issues, and provide structured judgments. 
We 

\subsection{Annotation Task}
For each sample, annotators are given:
\begin{itemize}[leftmargin=1.8em]
    \item \textbf{Website URL.} A link to the synthetic website instance.
    \item \textbf{Task instruction.} A natural-language description of the task the website is expected to support.
    \item \textbf{Task judge code.} A machine-checkable validator specification describing what constitutes task completion.
\end{itemize}

Annotators complete two sections:
\begin{itemize}[leftmargin=1.8em]
    \item \textbf{(A) Website Quality:} functional checks (feature-level checklist) and visual/appearance scoring (Likert scale).
    \item \textbf{(B) Task \& Judge Validity:} binary judgments on whether the task is executable and whether the judge correctly evaluates completion.
\end{itemize}

\subsection{(A) Website Quality}
Website Quality separates \emph{functional correctness} (whether features work) from \emph{visual realism} (how the site looks), to reduce confounding between broken UI behavior and poor styling.

\subsubsection{1) Core Functional Checks (Checklist)}
Annotators evaluate key interactive components of typical web services. For each functional area, annotators:
\begin{enumerate}[leftmargin=1.8em]
    \item Test the feature on the website (e.g., attempt signup if available).
    \item Select one status option.
    \item Optionally add a brief description of distinct issues encountered.
\end{enumerate}

Each functional check is a 3-way classification:
\begin{itemize}[leftmargin=1.8em]
    \item \textbf{Works correctly:} the feature behaves as expected without functional errors.
    \item \textbf{Broken / Not working as expected:} the feature fails, produces errors, or behaves incorrectly.
    \item \textbf{Not applicable:} the website does not include the feature (e.g., no login form exists).
\end{itemize}

The checklist covers:
\begin{itemize}[leftmargin=1.8em]
    \item \textbf{Signup / Registration} (if present).
    \item \textbf{Login} (if present; use provided test credentials if available).
    \item \textbf{Search functionality} (if a search bar or search UI exists).
    \item \textbf{Navigation \& links} (menus, primary links, buttons leading to other pages).
    \item \textbf{Forms \& submissions} (submit at least one form; verify success/error handling).
    \item \textbf{Filters / sorting / pagination} (if present in lists or search results).
\end{itemize}

\subsubsection{2) Visual / Appearance (Likert Scale)}
Annotators assess the overall realism and visual quality of the website independent of whether features function. Visual issues include (but are not limited to) misaligned elements, broken layouts, missing images/icons, inconsistent styling, or clearly unfinished design.

\paragraph{2.1 Overall Visual / Appearance Quality (1--5).}
Annotators select one rating based on the rubric below, counting \emph{distinct} visual issues rather than repeated instances:
\begin{itemize}[leftmargin=1.8em]
    \item \textbf{5 -- Excellent:} 0--2 very minor visual issues; overall layout resembles a real-world website.
    \item \textbf{4 -- Good:} 3--5 visual issues; minor inconsistencies but mostly realistic/professional.
    \item \textbf{3 -- Fair:} 6--8 visual issues; noticeable problems but still understandable and usable.
    \item \textbf{2 -- Poor:} 9--12 visual issues, or 1--2 severe visual failures (e.g., a key page is badly broken).
    \item \textbf{1 -- Very Poor:} more than 12 visual issues, or multiple severe layout failures (e.g., overlapping sections, unreadable text).
\end{itemize}

\subsection{(B) Task and Judge Validation}
This section evaluates whether the task is well-defined and executable on the site, and whether the judge correctly reflects true completion.

\subsubsection{Inputs}
Annotators are shown:
\begin{itemize}[leftmargin=1.8em]
    \item \textbf{Task Instruction:} the natural-language task to attempt on the website (e.g., ``search for a location and report rating and reviews'').
    \item \textbf{Task Judge Code:} a validation specification (e.g., a set of required substrings such as a target rating and review count, with an evaluation type).
\end{itemize}

\subsubsection{Binary Judgments}
\paragraph{1) Task Executability (Yes/No).}
Annotators judge whether the task can be completed using the website as described:
\begin{itemize}[leftmargin=1.8em]
    \item \textbf{Yes:} the task is doable with the website's available functionality and matches the instruction.
    \item \textbf{No:} the task is ambiguous, impossible, or depends on missing/non-functional components.
\end{itemize}

\paragraph{2) Judge Correctness (Yes/No).}
Annotators judge whether the validator accurately measures completion:
\begin{itemize}[leftmargin=1.8em]
    \item \textbf{Yes:} the judge accepts correct completions and rejects incorrect ones, consistent with the instruction.
    \item \textbf{No:} the judge produces false positives/negatives or does not match the instruction semantics.
\end{itemize}

\begin{figure*}
       \centering
       \includegraphics[width=1.0\linewidth]{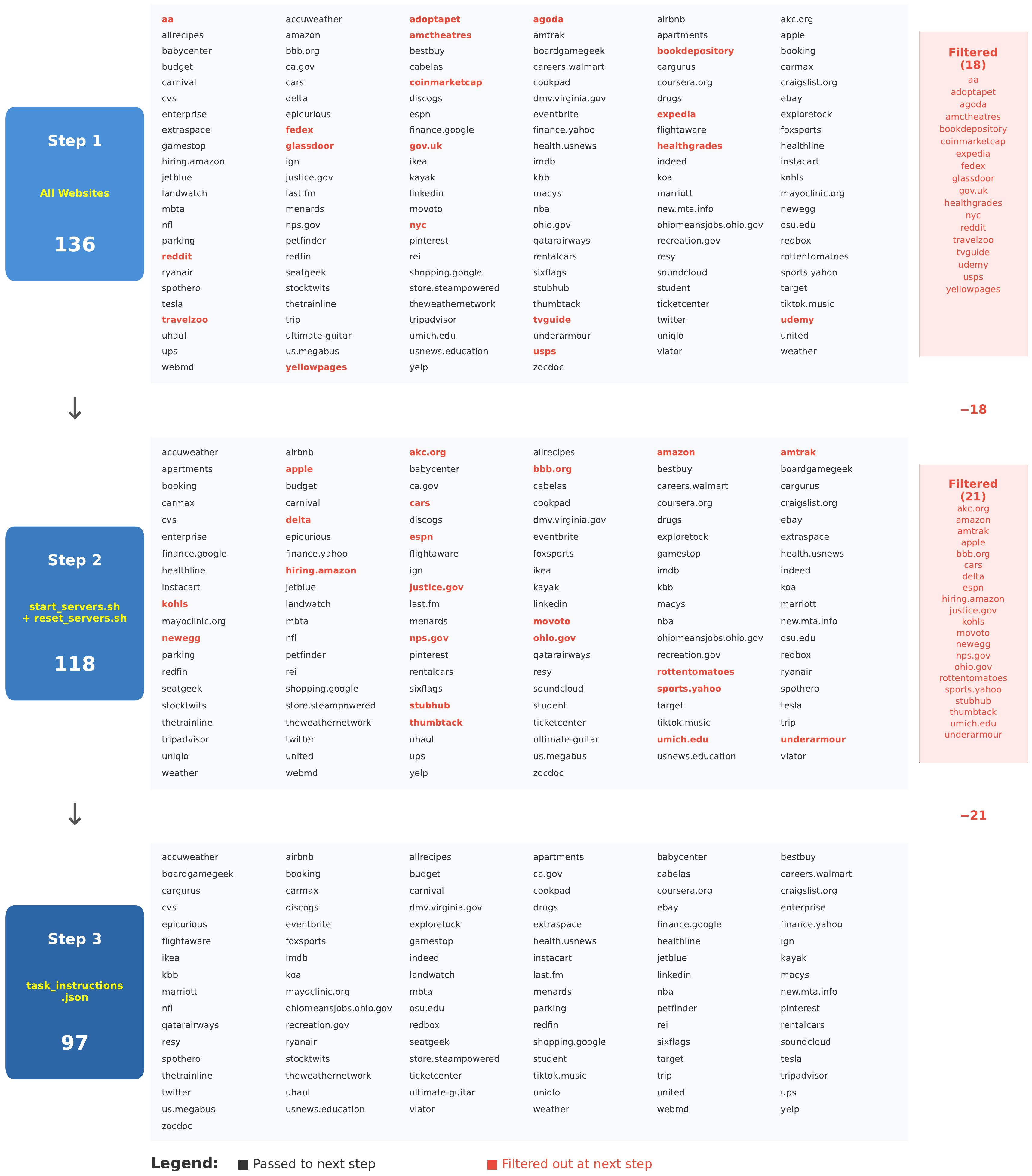}
       \caption{Website filtering flow for benchmark construction: 136 candidate websites are progressively filtered based on server script availability and task generation success, resulting in 97 validated benchmark websites.}
       \label{fig:website_dev}
\end{figure*}

\begin{longtable}{|p{2.2cm}|p{7cm}|p{4.5cm}|c|}
\caption{Example tasks and evaluation criteria from VeriEnv environments.}
\label{tab:example_tasks} \\
\hline
\textbf{Website} & \textbf{Task Instruction} & \textbf{Judge Criteria} & \textbf{Diff.} \\
\hline
\endfirsthead

\multicolumn{4}{c}{\tablename\ \thetable{} -- continued from previous page} \\
\hline
\textbf{Website} & \textbf{Task Instruction} & \textbf{Judge Criteria} & \textbf{Diff.} \\
\hline
\endhead

\hline
\multicolumn{4}{r}{{Continued on next page}} \\
\endfoot

\hline
\endlastfoot

adoptapet & Filter to cats and tell me how many results you get. & must\_include: ``2'' & easy \\
\hline
airbnb & On the home page, what categories are available in the category row? Please list all of them. & must\_include\_all: ``OMG!'', ``Lakefront'', ``Amazing pools'', ... (11 items) & easy \\
\hline
allrecipes & Open the Ingredients A--Z directory and click the letter A. Tell me the first five ingredient names listed under A. & must\_include: ``Apple Cider Vinegar'', ``Avocado'', ``albacore tuna'', ``alfalfa'', ``almond oil'' & easy \\
\hline
apartments & In Columbus, OH, sort listings by price (low to high). What is the first listing's name and minimum monthly price? & must\_include: ``The Charles at Bexley'', ``2938'' & medium \\
\hline
bestbuy & Sign in, save two specific products, open Saved Items and tell me how many total saved items. & exact\_match: ``2'' & easy \\
\hline
budget & Find the Red Ball Parking Garage location in New York, NY. Tell me the location code and hours. & must\_include: ``NYC1'', ``Mon-Sun 7:00 AM - 10:00 PM'' & easy \\
\hline
careers.walmart & Filter jobs to Drivers \& Transportation and pay type hourly. Tell me min and max pay. & must\_include: ``29'', ``36'' & easy \\
\hline
carmax & Go to favorites page and tell me how many cars are in favorites list. & must\_include: ``1'' & easy \\
\hline
coursera.org & Go to Resources, open 'Coursera Conference 2023', tell me the resource kind and summary. & must\_include: ``event'', ``Join leaders in higher education...'' & easy \\
\hline
cvs & Sort by price low to high, find first in-stock product, tell me name and price. & must\_include: ``Dudley Group Vitamin C Gummies'', ``\$9.99'' & easy \\
\hline
discogs & Open first item under 'Trending Releases'. Tell me release title and artist name. & must\_include: ``Open-architected maximized Local Area Network'', ``Chavez Trio'' & easy \\
\hline
drugs & Check interactions between 'Hydroxyzine' and 'Omeprazole'. Tell me severity and advice. & must\_include: ``minor'' & hard \\
\hline
epicurious & Open 'Gluten-Free Cinnamon Crumb Cake' recipe, tell me cuisine and rating. & must\_include: ``Korean'', ``3.0'' & easy \\
\hline
eventbrite & Check Tickets section. How many ticket types and cheapest price? & must\_include: ``2'', ``\$15'' & easy \\
\hline
exploretock & Open venue page for 'familiar formation'. Tell me address, city, state. & must\_include: ``220 Ash Street'', ``Chicago, IL'' & easy \\
\hline
extraspace & Search for Tampa, FL. Tell me star rating and review count. & must\_include: ``4.7'', ``13'' & easy \\
\hline
finance.google & Search for 'S\&P 500', tell me ticker symbol and current value. & must\_include: ``SPX'', fuzzy\_match: ``3746.49'' & easy \\
\hline
finance.yahoo & Find Microsoft's ticker symbol. & must\_include: ``MSFT'' & easy \\
\hline
foxsports & On Soccer league page, tell me first game's status. & must\_include: ``scheduled'' & easy \\
\hline
gamestop & Find highest priced featured product. Tell me name and price. & must\_include: ``Nintendo Switch OLED Model - White'' & easy \\
\hline
health.usnews & Search for 'keto'. Tell me type and title of first result. & must\_include: ``diet'', ``Keto Diet'' & easy \\
\hline
ign & Go to deals page, tell me exact title of first deal. & exact\_match: ``Outside goal official defense...'' & easy \\
\hline
instacart & In 'ALDI', find 'Basmati Rice - Family Size', tell me price. & must\_include: ``Basmati Rice - Family Size'' & easy \\
\hline
jetblue & Search one-way JFK to SFO for 2/1. Tell me lowest and highest price. & must\_include: ``320.00'', ``432.00'' & easy \\
\hline
linkedin & Log in, check Jobs alerts page, tell me how many alerts. & must\_include: ``3'' & easy \\
\hline
target & Search 'softwaves'. Tell me total results and first product title. & must\_include: ``32'', ``Bluetooth Speaker'' & easy \\
\hline
tesla & Filter Model 3 Used, sort by price. List first three prices. & must\_include: ``\$31,805'' & easy \\
\hline
thetrainline & From featured routes, find most expensive route (origin, destination, price). & must\_include: ``London'', ``Paris'', ``\$43.57'' & easy \\
\hline
uhaul & Create account, add items, remove one, tell me new subtotal. & must\_include: ``599'' & easy \\
\hline
ups & Estimate shipping from 90012 to 92101. Tell me cost and delivery days. & must\_include: ``9.82'', ``4'' & easy \\

\end{longtable}

\begin{figure*}[t]
\centering
\resizebox{1.0\textwidth}{!}{
\begin{tcolorbox}[
    colframe=darkgray,
    colback=gray!10,
    arc=2mm,
    boxrule=1pt,
    title={Prompts},
    fonttitle=\bfseries
]
\small\ttfamily
progress\_tracking\_and\_documentation: |\\
\hspace*{1em}Please use \texttt{todo.md} to track the progress of the implementation.\\
\hspace*{1em}Also, you have to document the implementation process in detail using markdown format.\\
\hspace*{1em}Use Linear service to track the progress of the implementation.\\[0.7em]

implementation\_prompt: |\\
\hspace*{1em}You are an expert programmer that implements the given website in full production quality.\\
\hspace*{1em}This is not a mockup website, so you need to implement all pages and features.\\[0.5em]

\hspace*{1em}Before implementing the website, inspect the ``landingpage.png'' and ``screenshot\_x.png'' files\\
\hspace*{1em}and write a detailed and structured description of the website in ``website\_description.md''.\\
\hspace*{1em}The implementation should be functionally and visually identical to the reference website.\\
\hspace*{1em}For the database, populate realistic data instead of only a few dummy entries.\\[0.5em]

\hspace*{1em}You may use any tools and libraries you want, but the final website must be fully deployable to production.\\
\hspace*{1em}For images, use assets from \texttt{https://images.unsplash.com/}.\\[0.5em]

\hspace*{1em}Also implement a Python SDK that can access the website's API, including authentication,\\
\hspace*{1em}database access, and other related functions.\\[0.5em]

\hspace*{1em}You have to run the website locally and check if it is working correctly.\\
\hspace*{1em}If the website is not working correctly, fix the bugs and run it locally again.\\
\hspace*{1em}Repeat this process until the website is working correctly.\\
\hspace*{1em}You can use Playwright MCP to check the website and obtain screenshots that show bugs or errors.\\[0.5em]

\hspace*{1em}After finishing the implementation, write a ``start\_servers.sh'' script to start the website locally.\\
\hspace*{1em}You need to run this script yourself. If it is not working, fix the bugs and rerun it.\\[0.5em]

\hspace*{1em}Lastly, implement a ``reset\_servers.sh'' script to reset the servers.\\
\hspace*{1em}Many instructions will modify the database state, so ``reset\_servers.sh'' should restore the DB\\
\hspace*{1em}to its initial state.\\[0.5em]

\hspace*{1em}Please use only the pre-assigned ports in ``ports.json''.\\
\hspace*{1em}Do not use any other ports and do not kill any other processes using other assigned ports.\\
\end{tcolorbox}
}
\caption{Prompt used for website reconstruction by the coding agent.}
\label{fig:prompt_website_reconstruction}
\end{figure*}
\begin{figure*}[t]
\centering
\resizebox{1.0\textwidth}{!}{
\begin{tcolorbox}[
    colframe=darkgray,
    colback=gray!10,
    arc=2mm,
    boxrule=1pt,
    title={Additional Prompts},
    fonttitle=\bfseries
]
\small\ttfamily
general\_prompt: |\\
\hspace*{1em}We are building a production-ready clone of the website.\\
\hspace*{1em}You need to refer to the ``landingpage.png'' and ``screenshot\_x.png'' files\\
\hspace*{1em}to implement a website that is fully functional and visually identical\\
\hspace*{1em}to the reference screenshots.\\
\hspace*{1em}In addition, implement a Python SDK that can be used to access the website's API,\\
\hspace*{1em}including the authentication process, database access, and related functions.\\[0.8em]

bug\_report\_prompt: |\\
\hspace*{1em}Check whether the server is running correctly.\\
\hspace*{1em}Refer to the ``start\_servers.sh'' file to start the server if needed.\\
\hspace*{1em}Ensure that ``reset\_servers.sh'' correctly resets the server state to its initial configuration.\\[0.5em]

\hspace*{1em}Analyze all functions and features of the website to identify bugs or errors.\\
\hspace*{1em}Use Playwright MCP to explore the website and capture screenshots that demonstrate\\
\hspace*{1em}the observed bugs or errors. Save screenshots in the ``screenshot'' directory\\
\hspace*{1em}(create it if it does not exist) using the filename format\\
\hspace*{1em}\texttt{bug\_report\_\{timestamp\}.png}.\\[0.5em]

\hspace*{1em}You may report multiple bugs. Visual similarity must also be considered by comparing\\
\hspace*{1em}against ``landingpage.png'' and ``screenshot\_x.png'' rather than images in the screenshot directory.\\
\hspace*{1em}Do not leave any placeholder buttons or links.\\[0.5em]

\hspace*{1em}Document each bug in a markdown file named\\
\hspace*{1em}\texttt{bug\_report\_\{timestamp\}.md}.\\[0.8em]

debug\_prompt: |\\
\hspace*{1em}After reporting all identified bugs, debug the website and fix the issues.\\
\hspace*{1em}You are a professional programmer responsible for patching the website.\\
\hspace*{1em}First, fetch the issue reports from GitHub.\\
\hspace*{1em}Test the website functionality thoroughly to ensure correct behavior.\\[0.5em]

\hspace*{1em}After completing the patches, provide a detailed debugging report.\\
\hspace*{1em}You may again use Playwright MCP to verify fixes and capture screenshots.\\
\hspace*{1em}Save screenshots in the ``screenshot'' directory using the filename format\\
\hspace*{1em}\texttt{debug\_\{timestamp\}.png}.\\
\hspace*{1em}Clearly explain the differences between the expected and actual results.\\[0.5em]

\hspace*{1em}Write the final debugging report to\\
\hspace*{1em}\texttt{debug\_\{timestamp, yyyy-mm-dd\_hh-mm-ss\}.md}.\\
\end{tcolorbox}
}
\caption{Prompts used for website implementation, bug reporting, and debugging by the coding agent.}
\label{fig:prompt_website_implementation_bug_reporting}
\end{figure*}
\begin{figure*}[t]
\centering
\resizebox{1.0\textwidth}{!}{
\begin{tcolorbox}[
    colframe=darkgray,
    colback=gray!10,
    arc=2mm,
    boxrule=1pt,
    title={Task and Judge Generation Prompt},
    fonttitle=\bfseries
]
\small\ttfamily
prompt: |\\
\hspace*{1em}Please generate 50 diverse task instructions that can be conducted within this website.\\
\hspace*{1em}Refer to the provided Python SDK when constructing the tasks.\\
\hspace*{1em}Each generated instruction must be fully validated using the Python SDK.\\[0.5em]

\hspace*{1em}The output file should be named \texttt{task\_instructions.json}.\\
\hspace*{1em}Avoid synthetic or unrealistic instructions.\\[0.5em]

\hspace*{1em}The output JSON file should contain a list of instruction objects.\\
\hspace*{1em}Each object must include the following fields:\\[0.3em]

\hspace*{2em}- \texttt{instruction}: A human-like task description (1--5 sentences).\\
\hspace*{2em}If authentication is required, the login process must be included.\\
\hspace*{2em}- \texttt{python sdk tool call}: SDK calls used to verify the task.\\
\hspace*{2em}- \texttt{tool call result}: The execution result of the SDK call.\\
\hspace*{2em}- \texttt{is\_valid}: Whether the instruction is valid.\\
\hspace*{2em}- \texttt{difficulty}: One of \texttt{easy}, \texttt{medium}, or \texttt{hard}.\\[0.5em]

\hspace*{1em}Difficulty definitions:\\
\hspace*{2em}- Easy: Simple browsing tasks with no authentication and minimal state changes.\\
\hspace*{2em}- Medium: Multi-step tasks with navigation and limited stateful actions.\\
\hspace*{2em}- Hard: Tasks requiring authentication and non-trivial state changes.\\[0.8em]

\hspace*{1em}judge\_for\_webagent format specification (IMPORTANT):\\
\hspace*{1em}The \texttt{judge\_for\_webagent} field must follow the exact JSON schema below.\\[0.4em]

\hspace*{1em}\{\\
\hspace*{2em}\texttt{"eval\_type"}: \texttt{"rinfo"} or \texttt{"rprog"},\\
\hspace*{2em}\texttt{"parse"}: \texttt{"json"} or \texttt{null},\\
\hspace*{2em}\texttt{"checks"}: [ \{ \texttt{"op"}, \texttt{"expected"}, \texttt{"path"} (optional) \} ]\\
\hspace*{1em}\}\\[0.6em]

\hspace*{1em}Supported operations for \texttt{checks}:\\
\hspace*{2em}\texttt{exact\_match}, \texttt{must\_include}, \texttt{fuzzy\_match}, \texttt{must\_include\_all}.\\[0.6em]

\hspace*{1em}Use \texttt{eval\_type = rinfo} for answer-based evaluation and\\
\hspace*{1em}\texttt{eval\_type = rprog} for programmatic verification (e.g., URL checks).\\
\hspace*{1em}Do not use deprecated string-based or natural-language judge formats.\\
\end{tcolorbox}
}
\caption{Prompt used to generate verifiable task instructions and executable judges using the Python SDK.}
\label{fig:prompt_task_judge}
\end{figure*}



\begin{figure*}[t]
    \centering
    \includegraphics[width=0.9\linewidth]{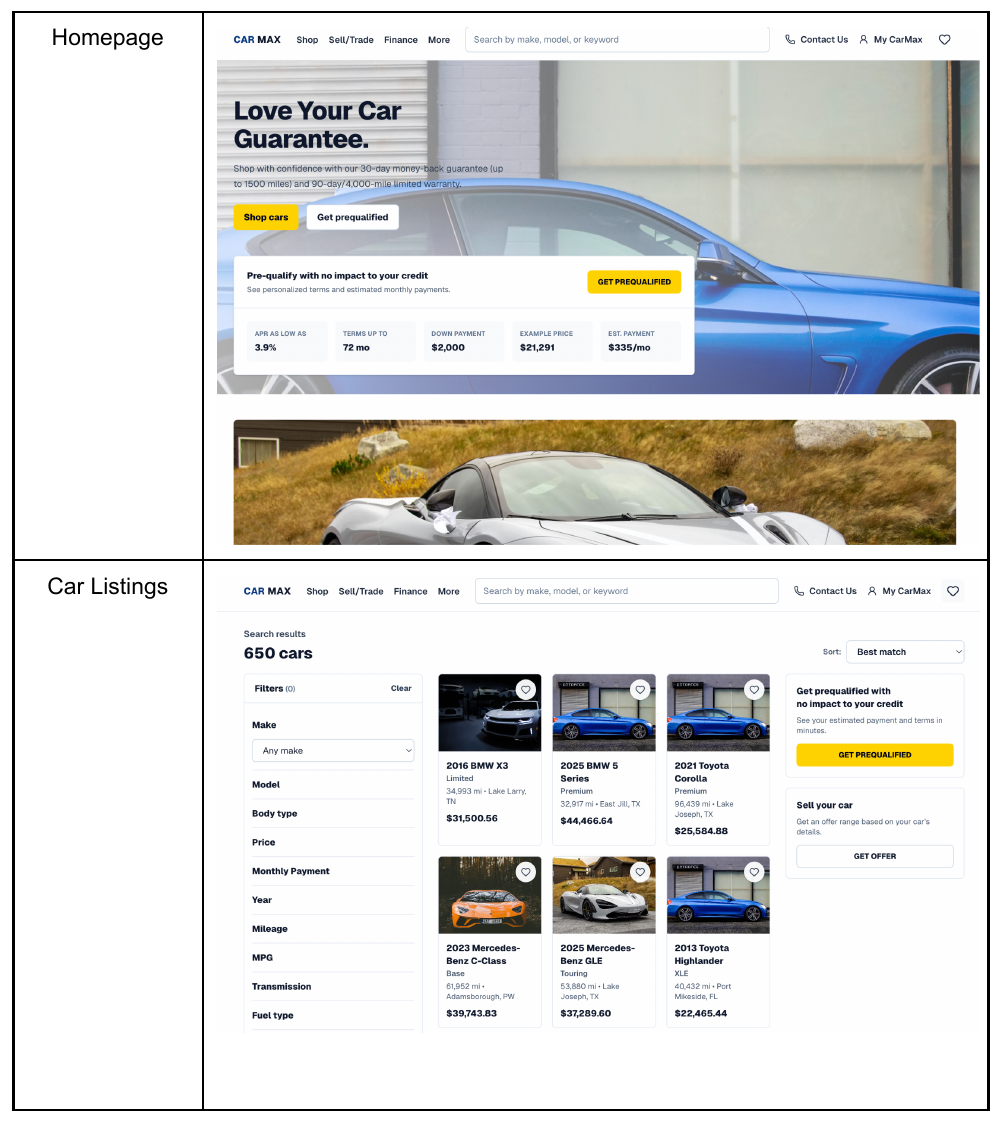}
    \caption{Screenshot from a cloned \textbf{CarMax} website (Part 1).}
    \label{fig:synth-web-1-1}
\end{figure*}

\begin{figure*}[t]
    \centering
    \includegraphics[width=0.9\linewidth]{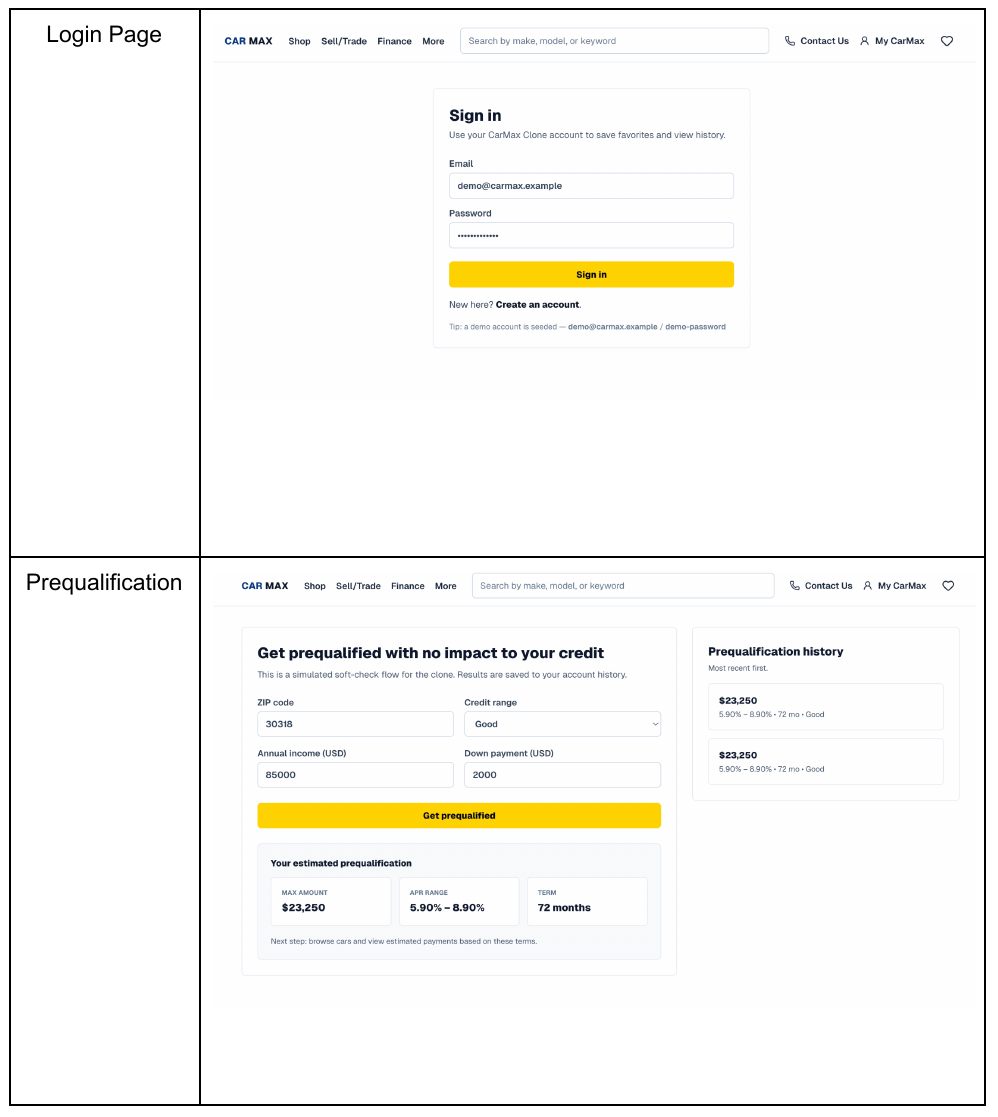}
    \caption{Screenshot from a cloned \textbf{CarMax} website (Part 2).}
    \label{fig:synth-web-1-2}
\end{figure*}

\begin{figure*}[t]
    \centering
    \includegraphics[width=0.9\linewidth]{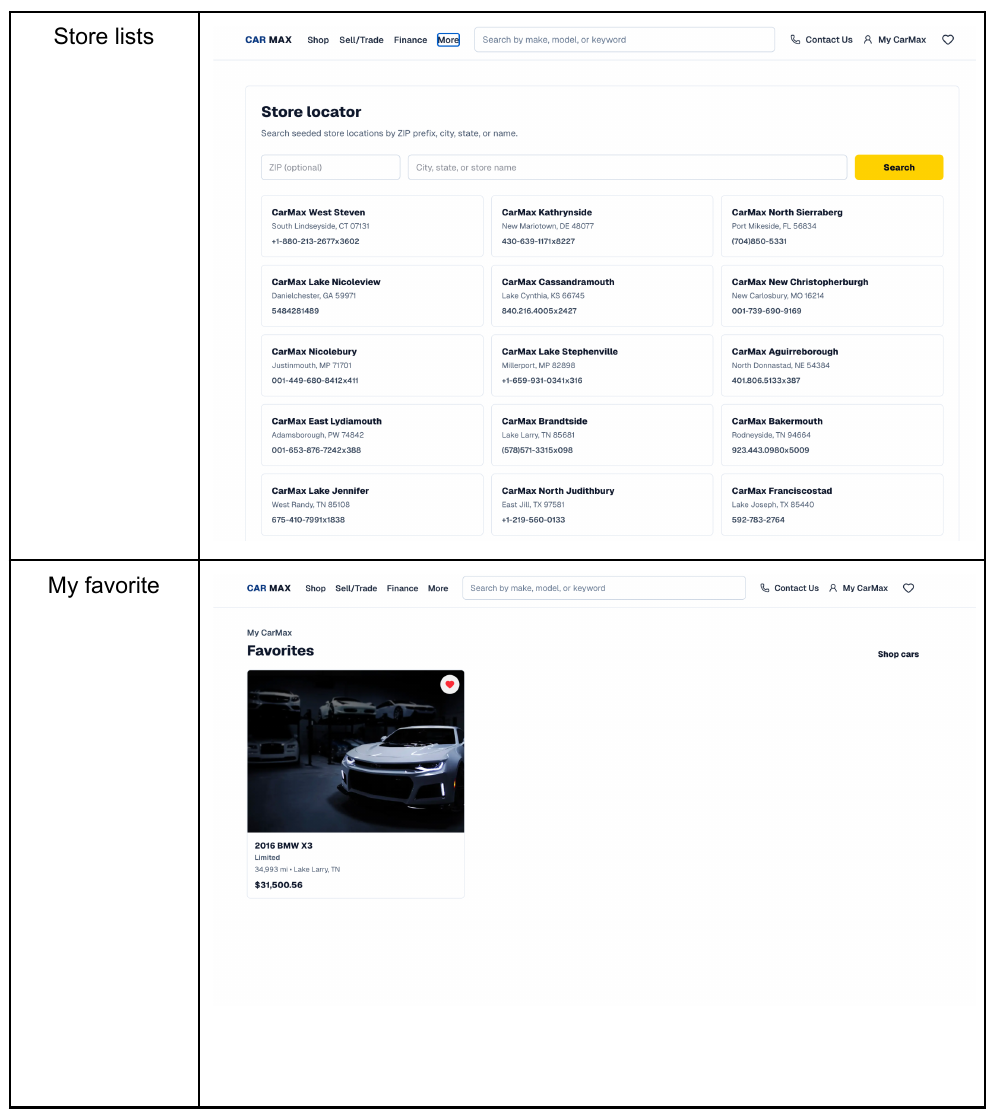}
    \caption{Screenshot from a cloned \textbf{CarMax} website (Part 3).}
    \label{fig:synth-web-1-3}
\end{figure*}

\begin{figure*}[t]
    \centering
    \includegraphics[width=0.9\linewidth]{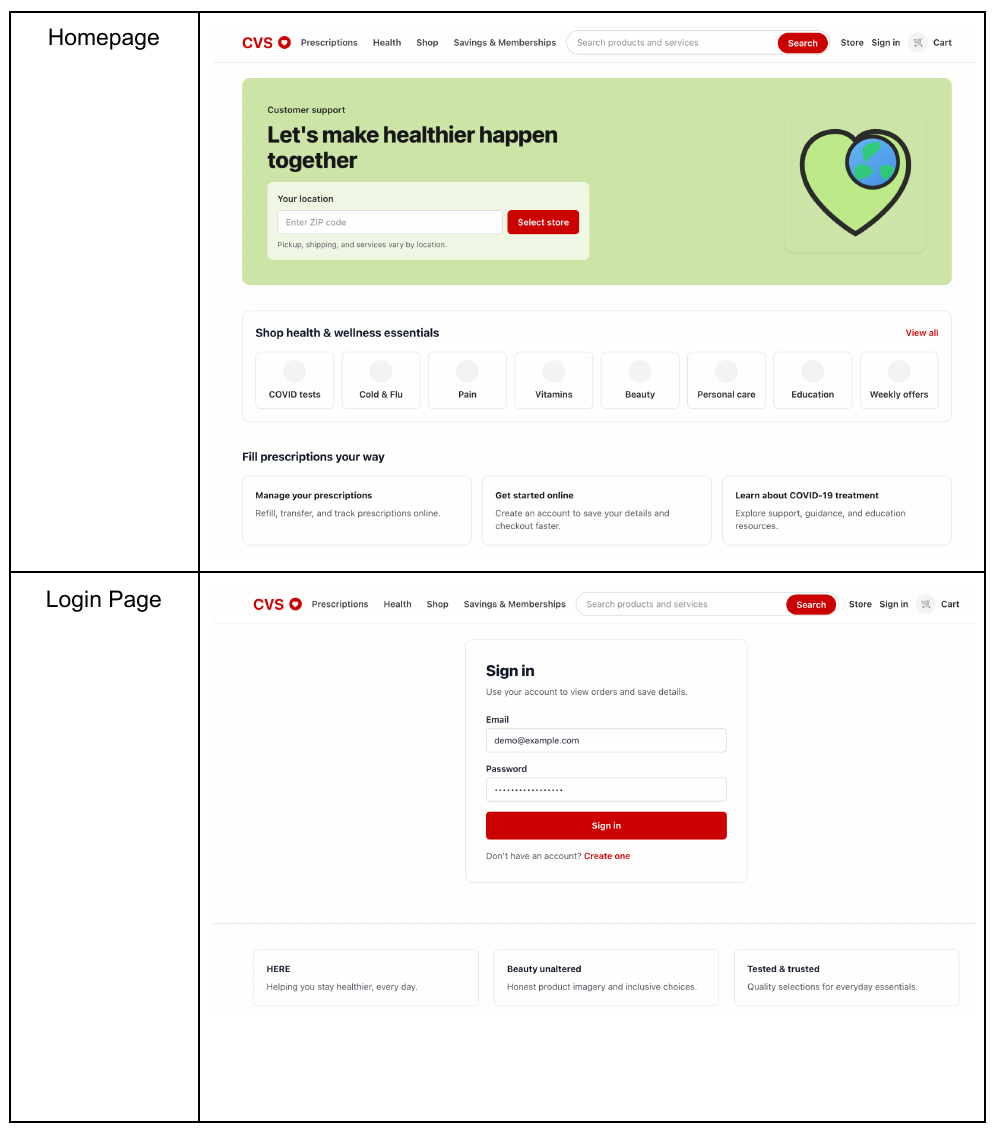}
    \caption{Screenshot from a cloned \textbf{CVS} website (Part 1).}
    \label{fig:synth-web-2-1}
\end{figure*}

\begin{figure*}[t]
    \centering
    \includegraphics[width=0.9\linewidth]{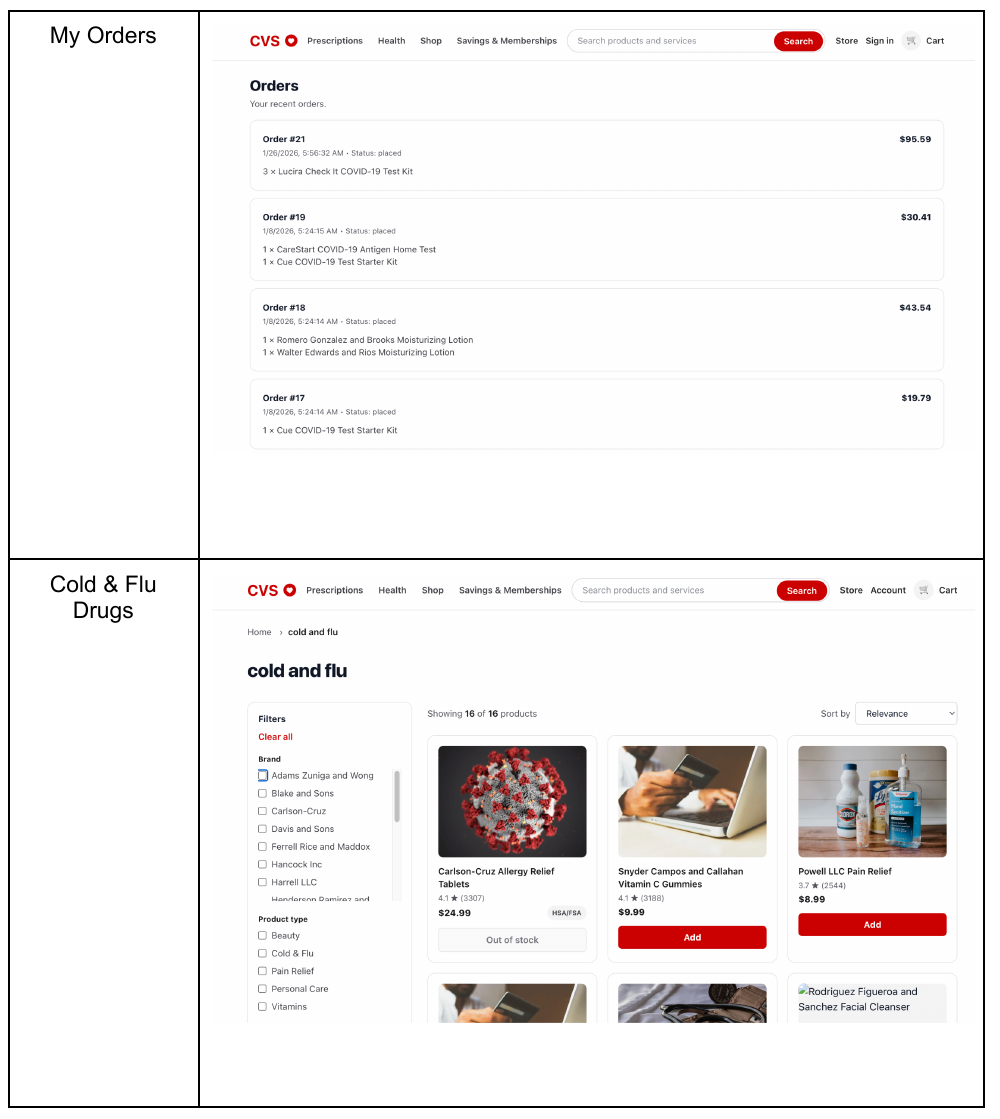}
    \caption{Screenshot from a cloned \textbf{CVS} website (Part 2).}
    \label{fig:synth-web-2-2}
\end{figure*}

\begin{figure*}[t]
    \centering
    \includegraphics[width=0.9\linewidth]{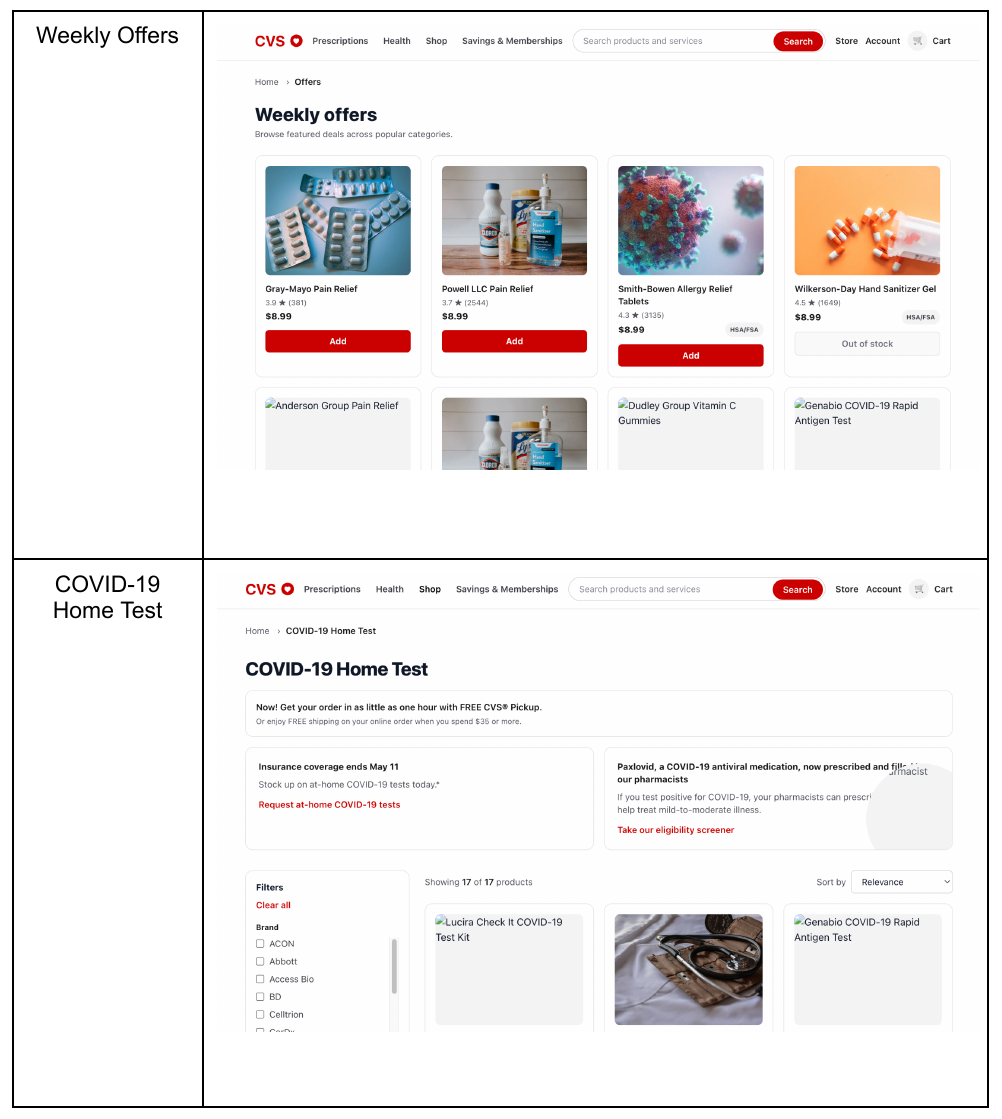}
    \caption{Screenshot from a cloned \textbf{CVS} website (Part 3).}
    \label{fig:synth-web-2-3}
\end{figure*}

\begin{figure*}[t]
    \centering
    \includegraphics[width=0.9\linewidth]{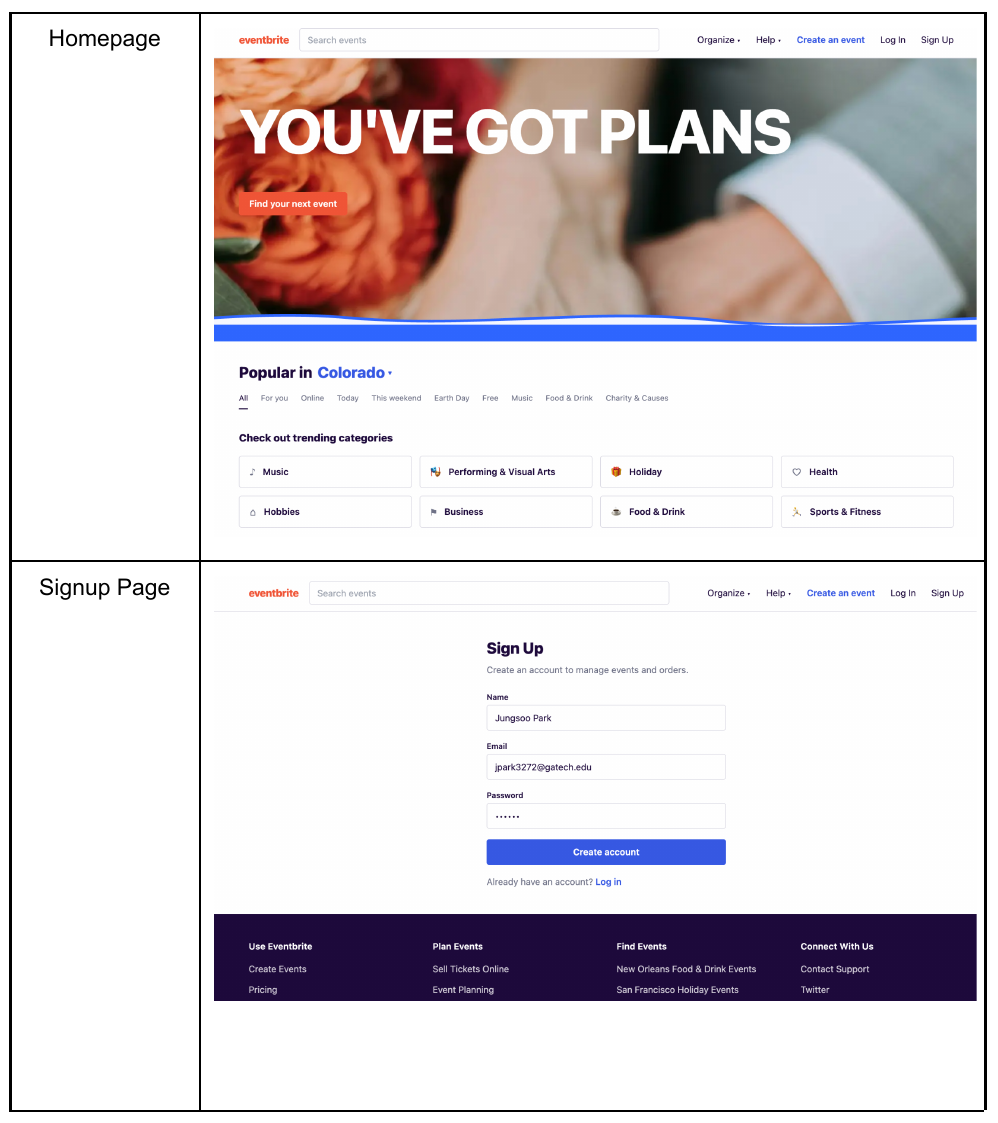}
    \caption{Screenshot from a cloned \textbf{eventbrite} website (Part 1).}
    \label{fig:synth-web-3-1}
\end{figure*}

\begin{figure*}[t]
    \centering
    \includegraphics[width=0.9\linewidth]{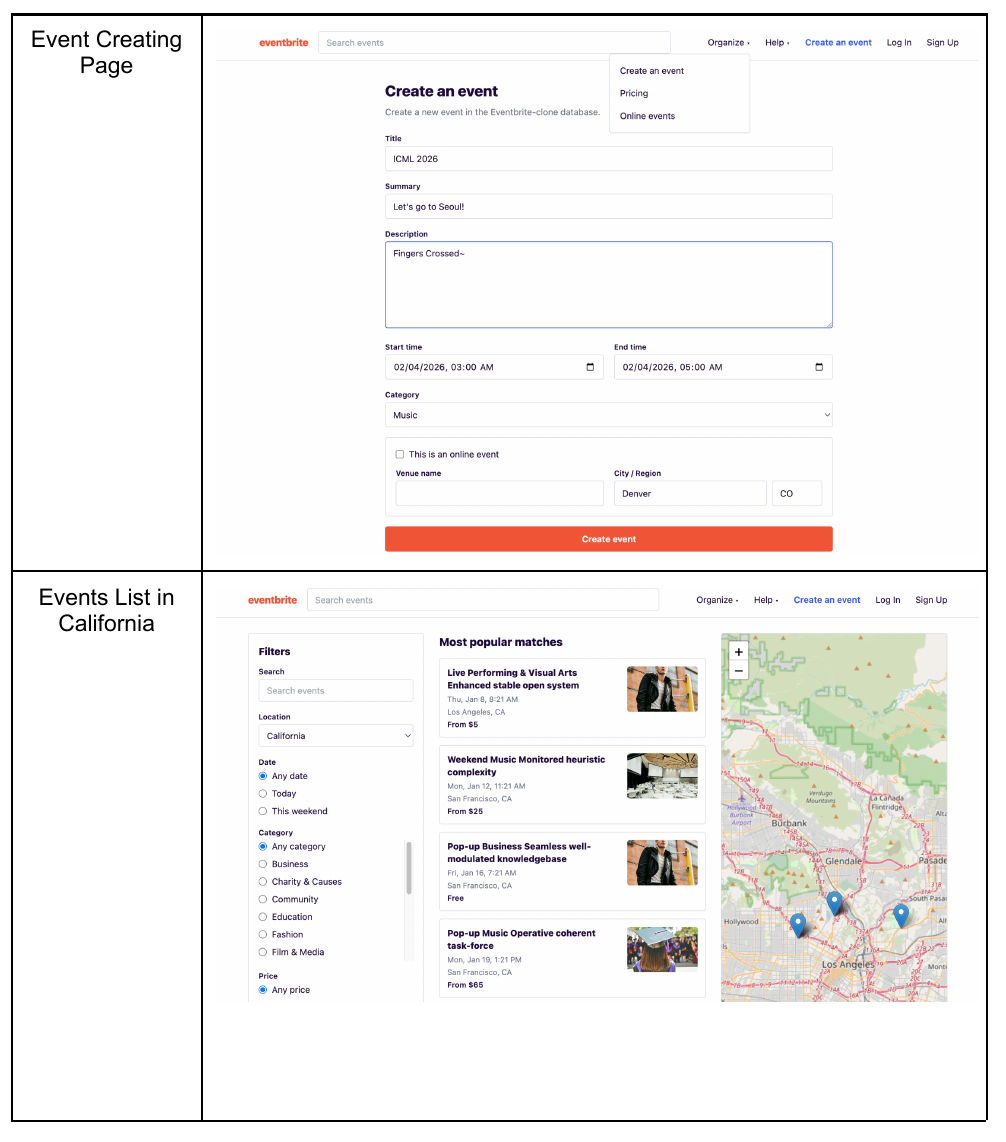}
    \caption{Screenshot from a cloned \textbf{eventbrite} website (Part 2).}
    \label{fig:synth-web-3-2}
\end{figure*}

\begin{figure*}[t]
    \centering
    \includegraphics[width=0.9\linewidth]{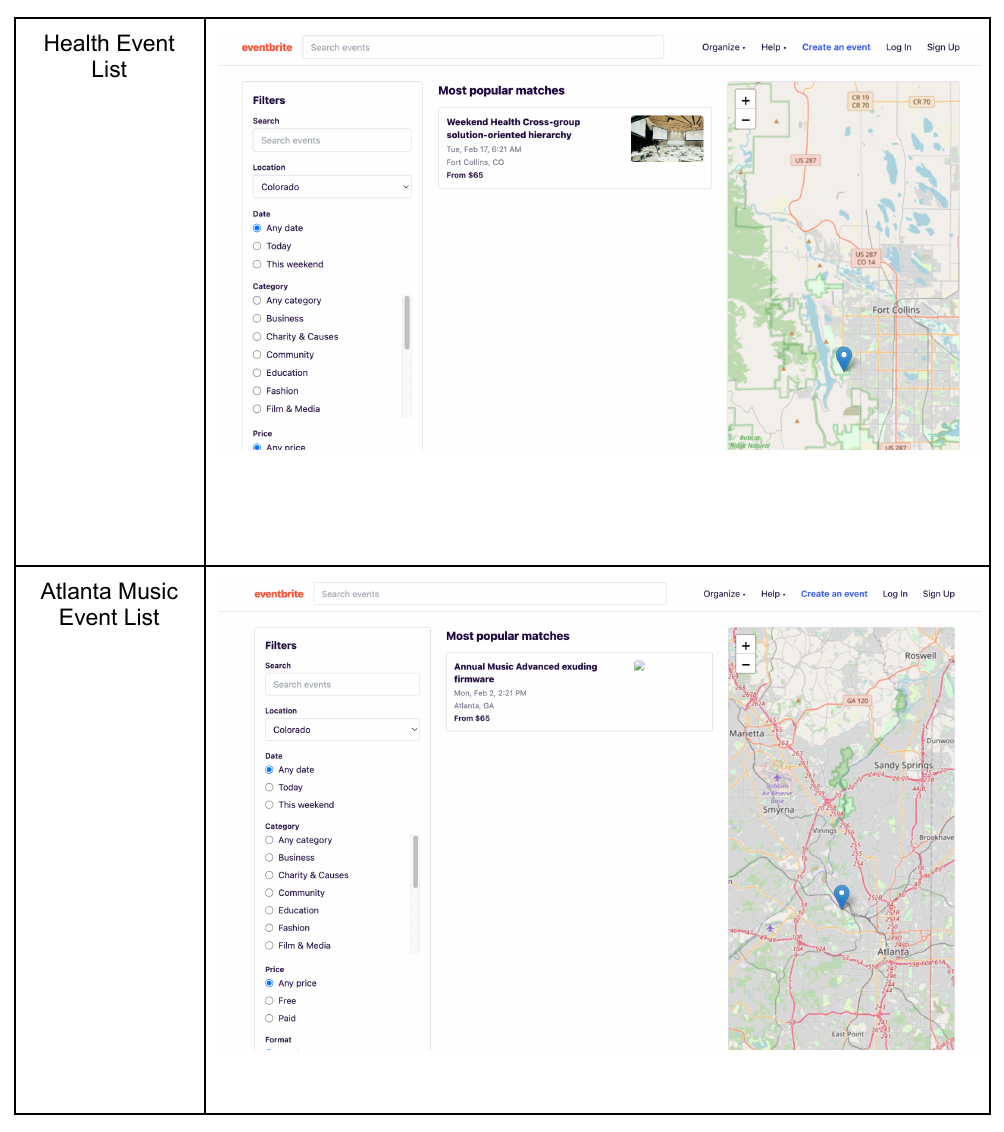}
    \caption{Screenshot from a cloned \textbf{eventbrite} website (Part 3).}
    \label{fig:synth-web-3-3}
\end{figure*}

\begin{figure*}[t]
    \centering
    \includegraphics[width=0.9\linewidth]{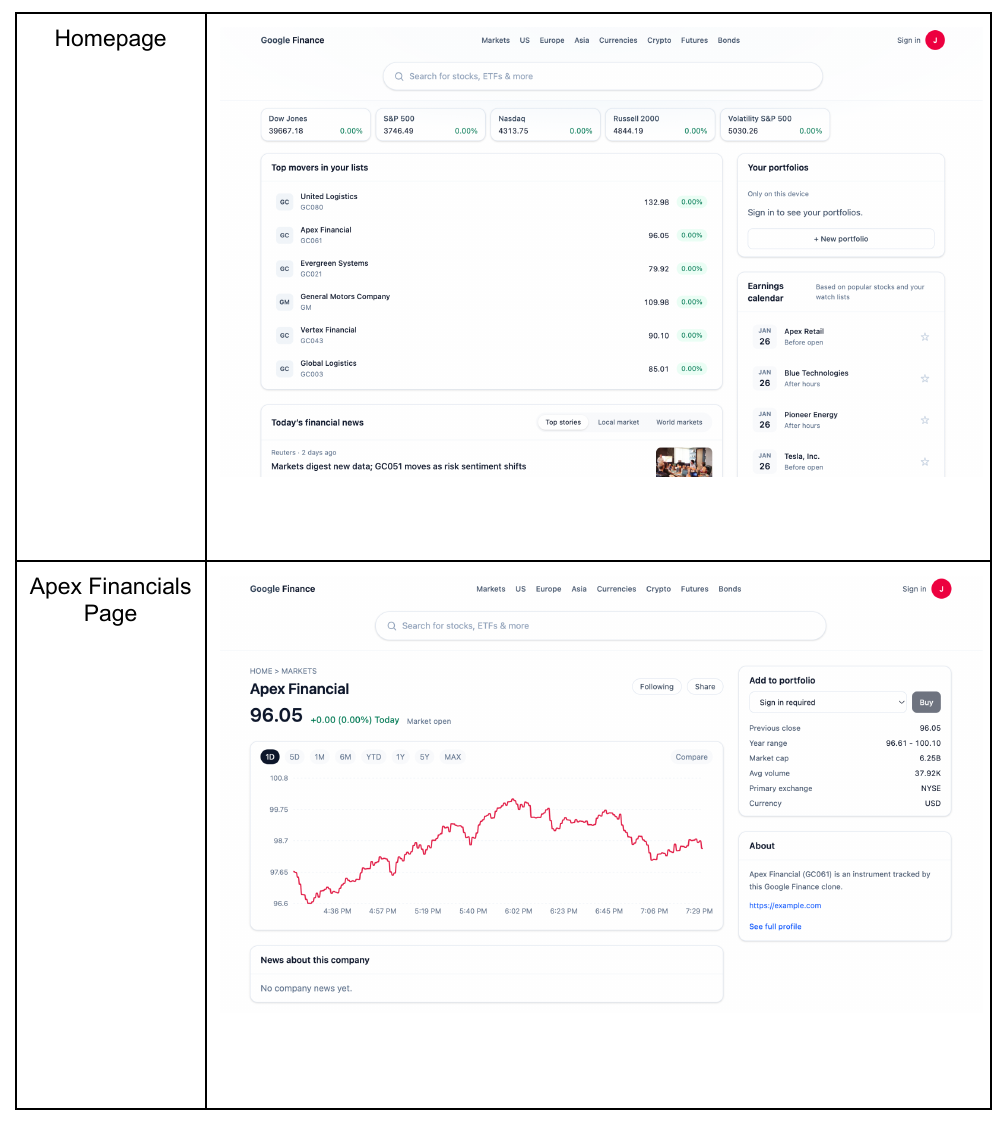}
    \caption{Screenshot from a cloned \textbf{Google Finance} website (Part 1).}
    \label{fig:synth-web-4-1}
\end{figure*}

\begin{figure*}[t]
    \centering
    \includegraphics[width=0.9\linewidth]{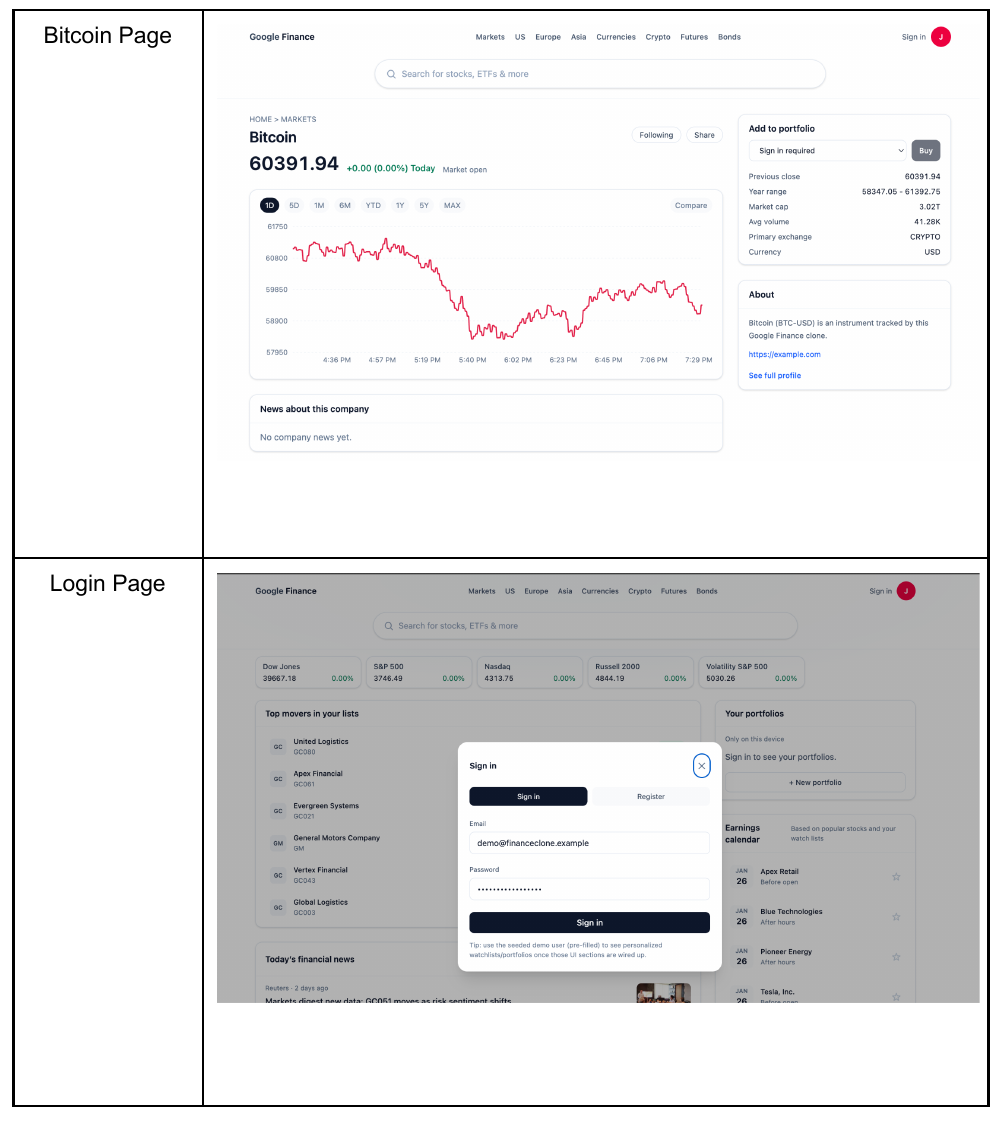}
    \caption{Screenshot from a cloned \textbf{Google Finance} website (Part 2).}
    \label{fig:synth-web-4-2}
\end{figure*}

\begin{figure*}[t]
    \centering
    \includegraphics[width=0.9\linewidth]{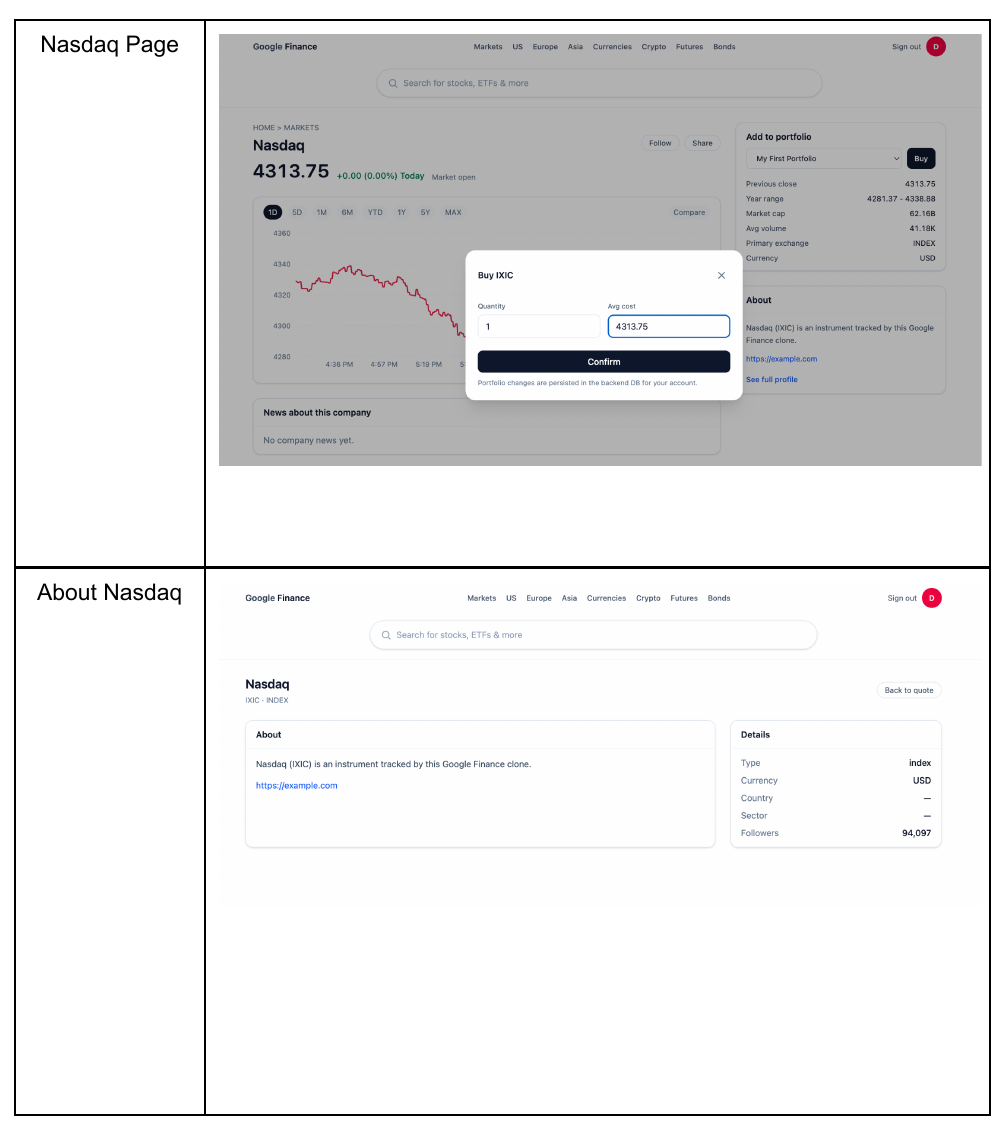}
    \caption{Screenshot from a cloned \textbf{Google Finance} website (Part 3).}
    \label{fig:synth-web-4-3}
\end{figure*}

\end{document}